\PassOptionsToPackage{table}{xcolor}  

\documentclass[sigconf, nonacm]{acmart}
\usepackage{enumitem}
\usepackage{subcaption}
\usepackage{multirow}
\usepackage{pifont}
\usepackage{listings}
\usepackage{xcolor}
\usepackage{algorithm}
\usepackage{algorithmic}
\usepackage{framed}  
\newcounter{insight}
\newenvironment{insightbox}{%
  \refstepcounter{insight}%
  \MakeFramed{\advance\hsize-\width \FrameRestore}%
  \small\noindent\textbf{Insight \theinsight:}~}%
  {\endMakeFramed}

\begin{document}

\title{Benchmarking Robust Aggregation in Gradient Data Marketplaces}

\author{Zeyu Song}
\email{zysong@psu.edu}
\orcid{0009-0005-9760-3493}
\affiliation{%
  \department{Department of Computer Science and Engineering}
  \institution{The Pennsylvania State University}
  \city{University Park}
  \state{Pennsylvania}
  \country{USA}
}

\author{Sainyam Galhotra}
\email{sg@cs.cornell.edu}
\orcid{0000-0003-2529-4036}
\affiliation{%
  \department{Department of Computer Science}
  \institution{Cornell University}
  \city{Ithaca}
    \state{New York}
    \country{USA}
}

\author{Shagufta Mehnaz}
\email{smehnaz@psu.edu}
\orcid{0000-0001-5850-4568}
\affiliation{%
  \department{Department of Computer Science and Engineering}
  \institution{The Pennsylvania State University}
  \city{University Park}
  \state{Pennsylvania}
  \country{USA}
}

\begin{abstract}

Institutions that cannot centralize or sell raw data increasingly exchange model updates instead: cross-silo federated learning is already in production among hospital networks and pharmaceutical consortia. A growing line of research targets the next step: marketplaces in which gradient updates are exchanged as paid, query-specific data products. No such marketplace is yet deployed at scale.
We formalize this emerging paradigm as a Gradient Data Marketplace (GDM), a demand-driven system in which data consumers issue learning tasks and distributed participants contribute query-specific gradient updates. Federated learning asks whether the model survives an attack; a marketplace must also ask who got paid.

Existing aggregation and evaluation protocols are designed to ensure model-level robustness, but fail to capture whether high-quality data is correctly identified and rewarded. In a GDM, selection directly determines payment, incentivizing participants to strategically optimize for inclusion rather than utility. To address this gap, we introduce a modular benchmarking framework for evaluating data selection and valuation mechanisms in gradient data marketplaces, separating \emph{mechanism integrity}---whether selection and payment reward true utility---from the \emph{model integrity} prior robustness work measures. We evaluate a standard baseline alongside 11 widely used aggregation mechanisms across vision, text, and tabular workloads using five real-world datasets.

Our results reveal a consistent pattern of adverse selection: honest high-variance contributions are rejected as outliers while strategically homogeneous malicious contributions are accepted and rewarded, distorting reward allocation even when model performance remains stable. These failures are not artifacts of our accounting: they persist under continuous, weight-proportional payment, compound across repeated market rounds as underpaid honest sellers exit, and worsen when honest sellers add differential-privacy noise, which the filters misread as contamination.

Our benchmark provides a foundation for designing incentive-aware, scalable, and auditable data selection mechanisms in emerging data marketplaces.

\end{abstract}



\received{17 April 2026}
\received[revised]{19 August 2026}
\received[accepted]{12 September 2026}

\maketitle

\section{Introduction}
\label{sec:intro}

Institutions that hold valuable data increasingly cannot centralize or sell it: privacy regulation such as HIPAA and GDPR turns on data minimization and purpose limitation, making raw-record transfer legally fraught even where demand for the underlying information is high. A task-specific derivative---composable with DP and secure aggregation~\cite{bonawitz2017practical}, which bound the inversion risk gradients otherwise carry~\cite{zhu2019deep}---is a categorically smaller disclosure than transferring records; and where sharing \emph{is} permitted, the seller retains the asset and can license it repeatedly, which transfer forecloses. Cross-silo federated learning (FL)~\cite{aggregation:fedavg} has become the production response, and is already operational at scale: the MELLODDY consortium ran federated training across ten competing pharmaceutical companies' proprietary compound libraries, aggregating partner gradients cryptographically at every iteration~\cite{oldenhof2023industry, heyndrickx2024melloddy}; FeatureCloud operates GDPR-compliant federated infrastructure across European clinical sites~\cite{matschinske2023featurecloud}; and full-stack federated learning has been deployed across four UK hospital groups to develop a COVID-19 screening test without centralizing patient data~\cite{soltan2024fullstack}. What every one of these operational systems lacks is an economic layer: participation is governed by consortium agreements and cost-sharing among pre-vetted peers, which works while participants are symmetric but breaks down as deployments scale to institutions with heterogeneous data quality, volume, and relevance---exactly the asymmetric setting a marketplace is designed for. A growing line of research targets precisely this next step: data marketplaces in which gradient updates are exchanged as paid, query-specific data products~\cite{dm:survey:zhang2024surveydatamarkets, liu2021dealer, gradient_market:chen2024federated, ddm:lu2024daved, gradient_market:li2022novel}, though no such marketplace is yet deployed at scale.

We formalize this setting as a \textbf{Gradient Data Marketplace (GDM)}, a demand-driven system in which a data consumer issues a learning task---analogous to a query---and distributed data providers respond with task-specific data derivatives (instantiated as gradient contributions). Unlike traditional data discovery systems, where static datasets are retrieved and integrated, GDMs operate over \emph{dynamically generated, query-conditioned data artifacts}, introducing new challenges in data selection, valuation, and integration.

As a running example, consider a hospital that holds a small labeled cohort of its own
and wishes to improve a diagnostic model. It issues a learning task; several peer clinics
respond with gradient updates computed on their local records; the broker selects a subset
and pays for it. Unknown to the broker, two of the ``clinics'' are one operator submitting under two identities, so a single participant collects twice. Every mechanism below---the query signal the hospital derives from its own cohort, the
filter that decides which updates are accepted, and the payment that follows from that
decision---acts on this scenario, and we return to it as each attack is introduced.

Existing FL robustness work evaluates whether the \emph{model} survives an attack---accuracy and attack success rate---and thereby conflates two distinct properties: \textbf{model integrity} (does the model survive?) and \textbf{mechanism integrity} (do selection and payment reward true utility?). In a marketplace the two come apart: a filter can preserve accuracy while paying camouflaged adversaries and rejecting honest but heterogeneous sellers, because \emph{data selection directly determines economic outcomes}. Participants are financially incentivized to maximize selection likelihood rather than data quality, creating a fundamental coupling between \emph{data management and economic incentives} that traditional FL evaluation protocols do not capture.

From a data systems perspective, GDMs raise four coupled challenges: \emph{selection integrity}, ensuring chosen contributions reflect true informational utility under heterogeneous distributions~\cite{zhao2018federated} rather than selection bias; \emph{incentive-compatible valuation}, aligning rewards with marginal contribution when participants optimize for selection likelihood rather than quality~\cite{fl:incentive:yang2020learning, fl:incentive:sim2023incentives}; \emph{scalable integration}, processing large candidate pools without $\mathcal{O}(N^2)$ overheads or discarding valuable data; and \emph{dual-sided strategic behavior}, since providers and consumers alike can shape selection outcomes and thus distort reward allocation.

\begin{figure*}[t]
    \centering
    \includegraphics[width=0.95\textwidth]{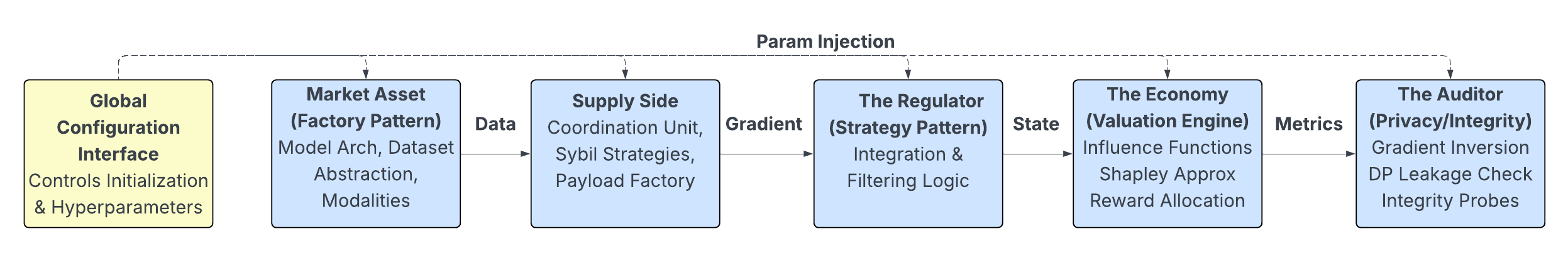} 
    \caption{\textbf{GDM Benchmark System Architecture.} 
    The framework is orchestrated by a central \textbf{Global Configuration} interface (Yellow) that injects hyperparameters into five decoupled modules (Blue). 
    \textbf{Solid arrows} show sequential data flow (from Data $\to$ Gradients $\to$ Metrics), while \textbf{dashed lines} show control flow (Parameter Injection). 
    This design structurally separates the \textit{Regulator} (\textbf{Data Integration/Filtering}) from the \textit{Economy} (\textbf{Data Valuation}), allowing for parallel auditing of mechanism integrity.}
    \label{fig:architecture}
\end{figure*}

These are fundamentally data management problems. A GDM is a query-driven data system: the buyer issues a learning task as a query, and the broker integrates query-conditioned data derivatives from distributed, untrusted sources. It inherits three classical concerns---\emph{integration} of heterogeneous sources under uncertainty~\cite{sheth1990federated}, \emph{provenance} tracking to support auditing~\cite{cheney2009provenance}, and \emph{query-based pricing} where selection determines payment~\cite{koutris2015query, balazinska2011data}---but diverges in one essential way: the underlying data is never directly accessible. Filters operate on derived signals rather than raw records, and the buyer's query signal $g_{ref}$ is itself an auditable but manipulable component of the mechanism. This creates failure modes that neither access-control-based database security nor standard FL robustness addresses: filters misclassify high-variance honest contributions as outliers, triggering adverse selection of exactly the data the buyer paid for, and buyers can weaponize the query signal to enforce monopolies without modifying any record. Evaluating these requires new abstractions---selection integrity, mechanism-aware valuation, query-signal auditing---alongside the infrastructure to instantiate them.

The pathway from today's consortia to deployed GDMs is concrete but gated on trust. The natural first deployers are existing federations themselves---consortium brokers among regulated peers---for whom production FL stacks and confidential-computing infrastructure already exist. Concretely, the transition is narrow: a MELLODDY-style consortium already computes per-partner gradients under secure aggregation every round, and already runs a trusted coordinator. Adding an economic layer means replacing the flat cost-sharing agreement with per-round selection and settlement---the filter that decides which partners' updates enter the aggregate becomes the thing that decides who is paid. No new distributed infrastructure is required; what changes is that a filtering decision, previously invisible, acquires a price. What is missing is not systems machinery but assurance about the mechanism: no operator will attach payments to a selection pipeline that camouflaged contributors can game, and no honest seller will remain in a market that systematically rejects heterogeneous data. \emph{Auditing mechanism integrity is therefore a precondition for deployment---and existing FL benchmarks, built to measure model survival, lack the metrics to provide it.}
To bridge this gap, we introduce a modular benchmark framework for evaluating data selection and integration in incentive-driven markets (Figure~\ref{fig:architecture}), whose decoupled supply, filtering, valuation, and auditing modules let researchers isolate a mechanism's impact on both model utility and economic integrity.

\vspace{1mm}\noindent\textbf{GDM Setup and Evaluation Focus.}
To operationalize our benchmark, we align our scope with the practical realities of data marketplaces, differentiating our setup from traditional FL in two key ways. First, we focus strictly on a central broker utilizing explicit \textit{integration operators}---including per-seller rejection, coordinate-wise clipping, and reference-weighted aggregation---rather than peer-to-peer protocols. 
Second, we exclude peer-to-peer cross-validation, which in a revenue-sharing economy incentivizes cartels and artificial downvoting (Section~\ref{sec:related_work}).

\noindent\textbf{Our Contributions.}
To bridge theoretical FL security and practical GDM viability, we introduce a benchmark framework with the following contributions:
\begin{enumerate}[leftmargin=*]
   \item \textbf{Modular Sandbox for Defense Prototyping:} An open-source pipeline\footnote{\url{https://github.com/szyszyzys/dgm-bench.git}} that decouples the supply, integration, and valuation layers, letting researchers prototype incentive-aware integration operators without rebuilding distributed infrastructure.
    
\item \textbf{Dual-Sided Threat Modeling for Data Integrity:} 
We identify vulnerabilities unique to the interaction between 
financially motivated supply and demand, modeling strategic 
behavior on both sides through representative attack strategies:
\begin{itemize}[nosep]
    \item \textit{Supply-Side (Rational Seller):} Coordinated 
    \textbf{Sybil Drowning} attacks that exploit clustering logic 
    to spoof consensus, displacing honest competitors.
    \item \textit{Demand-Side (Adversarial Buyer):} \textbf{Geometric 
    Pivoting} and \textbf{Dynamic Instability} attacks that weaponize the query signal ($g_{ref}$) to enforce monopolies or systematically exclude sellers, an adversary that root-of-trust designs assume honest~\cite{fl:aggregation_method:li2023martfl}.
\end{itemize}
    
    \item \textbf{Marketplace-Centric Systems Evaluation:} We augment standard robustness metrics with explicit economic and system dimensions. We introduce \textit{Malicious Selection Rate (MSR)} and \textit{Benign Selection Rate (BSR)} to quantify payout theft and honest data rejection, alongside \textit{Cost-of-Convergence (CoC)}, \textit{Broker Latency}, and \textit{Peak Memory Overhead}. This allows our benchmark to distinguish between \textit{Model Integrity} (did the model survive?) and \textit{Mechanism Integrity} (did the attacker get paid efficiently?)---a dimension prior FL benchmarks do not evaluate.

    \item \textbf{Fundamental Limitations of Geometric Filtering:} We empirically demonstrate that current software-only integration filters face a structural limit in distinguishing benign data heterogeneity from adversarial poisoning. We quantify this phenomenon as \textit{Adverse Selection}---where the GDM inadvertently bankrupts high-variance, honest sellers while actively subsidizing malicious actors. We conclude that robust GDMs cannot be secured by spatial algorithms alone, requiring a fundamental shift toward economic deterrence or hardware-rooted trust. 

\end{enumerate}
\section{Related Work}
\label{sec:related_work}

\begin{table*}[t]
\footnotesize
\centering
\caption{\textbf{Comparison with Existing Federated Learning Benchmarks.} While prior work focuses on optimization efficiency, scalability, or model security, our framework is the first to quantify mechanism integrity (economic security) against rational adversaries in a dual-sided data derivative marketplace.}
\label{tab:benchmark_comparison}
\begin{tabular}{lcccccc}
\toprule
\textbf{Benchmark Feature} & \textbf{LEAF}~\cite{fl:benchmark:caldas2018leaf} & \textbf{FedML}~\cite{fl:benchmark:he2020fedml} & \textbf{FedScale}~\cite{lai2022fedscale} & \textbf{OpenDataVal}~\cite{jiang2023opendataval} & \textbf{OARF}~\cite{hu2022oarf} & \textbf{Our Work (GDM)} \\
\midrule
\textbf{Primary Focus} & Data Heterogeneity & Distributed ML Dev. & System Scalability & Static Data Valuation & Model Security & \textbf{Mechanism Integrity} \\
\textbf{Threat Model} & N/A & Seller Poisoning & Stragglers/Failures & N/A & Seller Poisoning & \textbf{Dual-Sided (Buyer+Seller)} \\
\textbf{Integrity Metric} & Accuracy & Accuracy / Speed & Throughput & Shapley Error & ASR (Attack Success) & \textbf{MSR, BSR, CoC} \\
\textbf{Adversary Goal} & N/A & Model Destruction & N/A & N/A & Model Destruction & \textbf{Profit Maximization} \\
\textbf{Participation} & Transient & Transient & Transient & Static & Transient & \textbf{Continuous/Persistent} \\
\textbf{Root of Trust} & N/A & Server (Honest) & Server (Honest) & Ground Truth & Server (Honest) & \textbf{Query Signal ($g_{ref}$)} \\
\bottomrule
\end{tabular}%
\end{table*}

Our work bridges classical distributed data management~\cite{mottin2017new, sheth1990federated} and decentralized data economics. As summarized in Table~\ref{tab:benchmark_comparison}, existing benchmarks prioritize system optimization or model security. We instead audit these operators strictly as \textit{transaction verification engines}, introducing dual-sided economic threat models to evaluate data integration under rational, profit-seeking behavior.

\noindent\textbf{Robustness and Benchmarking.}
Standard benchmarks (e.g., FedML \cite{fl:benchmark:he2020fedml}) evaluate filters including distance-based aggregators (Multi-Krum~\cite{fl:agg:krum:blanchard2017machine}, Trimmed Mean~\cite{fl:agg:trimmedmean:yin2018byzantine}, Bulyan~\cite{fl:agg:guerraoui2018hidden}), reference-based filters (FLTrust~\cite{fl:aggregation_method:cao2020fltrust}), and backdoor-specific defenses (FLAME~\cite{fl:agg:nguyen2022flame}, DeepSight~\cite{fl:agg:rieger2022deepsight}) on global loss.
Security benchmarks (e.g., OARF~\cite{hu2022oarf}) optimize for \textit{Model Integrity} (minimizing Attack Success Rate). 
From a marketplace perspective, indiscriminately discarding 50\% of paid data achieves high Model Integrity but triggers a catastrophic \textit{Mechanism Integrity} failure. We distinguish our framework by explicitly auditing this economic cost, proving current mechanisms allow adversaries to steal rewards while triggering the adverse selection of honest data.

\noindent\textbf{Backdoor Attacks in Federated Learning.}
We adopt the standard backdoor threat model~\cite{attack:fl:backdoor:bagdasaryan2020backdoor}, in which adversaries embed trigger-conditioned behavior while preserving benign-task accuracy. Backdoors expose selection-as-payment failures that untargeted poisoning cannot: their stealth lets attackers maintain high similarity scores and collect rewards.

\noindent\textbf{Data Marketplaces \& Pricing.}
Classical databases study query-based pricing to prevent consumers 
from reconstructing premium data~\cite{koutris2015query, 
balazinska2011data}, and modern data-market designs extend this with 
algorithmic allocation and fairness 
guarantees~\cite{market:agarwal2019marketplace, dm:survey:zhang2024surveydatamarkets,castro2023data}. 
GDMs face the reverse vulnerability: sellers submitting plausible 
noise to steal payouts, a form of free-riding previously studied in 
cooperative FL~\cite{fairness:lin2019free, fairness:chen2024fair}. 
Furthermore, current query-driven FL protocols (e.g., 
MartFL~\cite{fl:aggregation_method:li2023martfl}) assume honest 
buyers, ignoring the demand-side threat where a malicious buyer 
weaponizes the query signal ($g'_{ref}$) to enforce artificial 
monopolies---the natural counterpart, in gradient marketplaces, to 
strategic-buyer threats previously studied in raw-data 
markets~\cite{market:castrofernandez2022protecting}.

\noindent\textbf{Data Valuation.}
Exact Shapley Values~\cite{valuation:shapley:ghorbani2019data} ensure
fair attribution but remain computationally prohibitive
($\mathcal{O}(N!)$). Tractable approximations reduce this cost
through nearest-neighbor
structure~\cite{valuation:jia2019knnshap} or kernel-based local
surrogates~\cite{valuation:shapley:lundberg2017unified},
while alternative indices trade axiomatic guarantees for robustness or tractability, including leave-one-out and influence functions~\cite{valuation:influence:koh2017understanding, valuation:ilyas2022datamodels}, the Banzhaf value~\cite{wang2023data}, and the Core~\cite{yan2021if}. 
Proposed GDM designs nonetheless default to cheap geometric heuristics like cosine similarity~\cite{fl:aggregation_method:li2023martfl}.
Our benchmark reveals that strategic adversaries easily exploit this
efficiency-fairness trade-off, gaming similarity metrics to capture
revenue while actively degrading the aggregated data asset.

\noindent\textbf{Privacy-Priced Marketplaces.}
A complementary line builds the market \emph{around} differential privacy, pricing locally perturbed gradients through auctions~\cite{zheng2022fl} or Stackelberg games that derive profit-maximizing~\cite{sun2024profit} and socially optimal~\cite{sun2024socially} pricing. These designs optimize \emph{what privacy costs}, assuming contributions are aggregated as submitted. Our benchmark audits the selection layer they leave untouched and exposes a tension: the calibrated DP noise such markets compensate is what distance-based filters misclassify as contamination, suppressing honest-seller selection (the \emph{Privacy Tax}, Section~\ref{sec:results_privacy_tax}). A robust GDM thus under-selects the very sellers a privacy-priced market is designed to pay; reconciling the two is an open problem our framework is positioned to support.

\noindent\textbf{Fairness and Incentive Compatibility.}
Parallel work targets free-riding and fair contribution via 
game-theoretic incentives~\cite{fairness:chen2024fair, fl:incentive:sim2023incentives} and 
contribution-based valuation in FL~\cite{fairness:wang2020principled}. 
We explicitly exclude peer-to-peer cross-validation; in a 
revenue-sharing economy, peer-grading incentivizes cartels and 
artificial downvoting. Instead, we focus on the central broker, 
auditing whether distance-based filters inadvertently penalize 
high-variance sellers while rewarding artificial conformity.

\noindent\textbf{Verifiable Systems \& Provenance.}
TEEs (e.g., Intel SGX) secure processing against 
tampering~\cite{mo2021ppfl, burkhalter2023rofl}, with newer 
generations (Intel TDX, AMD SEV-SNP) offering GB-scale secure 
memory~\cite{tee:cheng2024intel}. Complementary primitives include 
secure aggregation~\cite{bonawitz2017practical} and database 
provenance tracking~\cite{cheney2009provenance}. However, none of 
these resolve the validation paradox for real-valued data derivatives: if a buyer provides a biased reference ($g'_{ref}$), the hardware faithfully computes a rigged valuation. We benchmark the systemic failures of these software algorithms to highlight exactly where computational guarantees and economic security currently diverge.
\section{The GDM Benchmark Framework}
\label{sec:system_framework}

Evaluating a Gradient Data Marketplace (GDM) requires shifting focus from standard machine learning convergence to the efficiency and security of a distributed data pipeline. Unlike static datasets, a GDM operates as a dynamic, query-driven data broker. To capture this, we propose a compositional data pipeline framework that strictly decouples data integration, quality filtering, asynchronous auditing, and fair pricing, treating seller retention and runtime extensibility as design constraints rather than afterthoughts. Throughout, we use \emph{gradient contribution} and \emph{data derivative} interchangeably to refer to a seller's submitted update, and \emph{Broker} to refer to the central Integration Engine.

\subsection{Marketplace-Centric Metrics}
\label{sec:metric_taxonomy}
Standard ML metrics fail to capture the system overhead and financial incentives of a GDM. We therefore introduce a multi-dimensional scorecard (Table~\ref{tab:metrics}) explicitly decoupling System Performance, Mechanism Integrity, and Model Integrity. To formalize our economic metrics, let $\mathcal{S}_{adv}$ denote the set of adversarial sellers, and $\mathcal{S}_{sel}$ the subset of sellers who pass the quality filter to receive payment.
Under non-binary payment models, these selection-based metrics generalize directly: economic capture becomes the adversarial \emph{revenue share}---the fraction of total payout received by $\mathcal{S}_{adv}$ (Section~\ref{sec:valuation}).

\begin{table}[htbp]
\centering
\small
\renewcommand{\arraystretch}{1.2}
\caption{\textbf{GDM Metric Taxonomy.} We explicitly decouple System Performance (broker overhead) from Mechanism Integrity (economic governance) and Model Integrity (data product quality).}
\label{tab:metrics}
\begin{tabular}{l|p{0.75\columnwidth}}
\hline \hline
\multicolumn{1}{l}{\textbf{Metric}} & \textbf{Definition \& GDM Implication} \\
\hline
\multicolumn{2}{c}{\cellcolor{gray!10}System Performance (Broker Overhead)} \\
\hline
\textbf{T\textsubscript{round}} & Broker Latency. Wall-clock time required for the broker to execute integration filters per round. \\
\textbf{M\textsubscript{peak}} & Peak Broker Memory. Maximum RAM/VRAM footprint required by the integration operator to maintain state. \\
\hline
\multicolumn{2}{c}{\cellcolor{gray!10}Mechanism Integrity (Economic Security)} \\
\hline
\textbf{MSR} & Malicious Selection Rate. The fraction of malicious sellers successfully selected ($|\mathcal{S}_{sel} \cap \mathcal{S}_{adv}| / |\mathcal{S}_{adv}|$). Indicates revenue theft. \\
\textbf{BSR} & Benign Selection Rate. The fraction of available honest sellers selected. Low BSR indicates adverse selection and discarded data diversity. \\
\textbf{CoC} & Cost of Convergence. Total upload volume (MB) consumed to reach target accuracy $\tau_{acc}$, or over the full run if $\tau_{acc}$ is not reached. A proxy for Return on Investment (ROI) on data acquisition. \\
\hline
\multicolumn{2}{c}{\cellcolor{gray!10}\textbf{Model Integrity (Data Product Quality)}} \\
\hline
\textbf{ASR} & Attack Success Rate. The percentage of triggers misclassified to the target label. \\
\textbf{ACC} & \textbf{Accuracy.} Predictive accuracy of the final aggregated model on the buyer's task. \\
\hline
\end{tabular}
\end{table}

\noindent\textbf{Separating Revenue Theft from Model Damage.}
Model-centric metrics hide an economic reality: parasitic adversaries maximize MSR (revenue theft) without inflating ASR (model damage), while overzealous filters achieve low ASR at low BSR, starving legitimate participants and wasting broker budget.

\subsection{Modular System Architecture}
\label{sec:system_architecture}

\begin{figure}[t]
    \centering
    \includegraphics[width=\columnwidth]{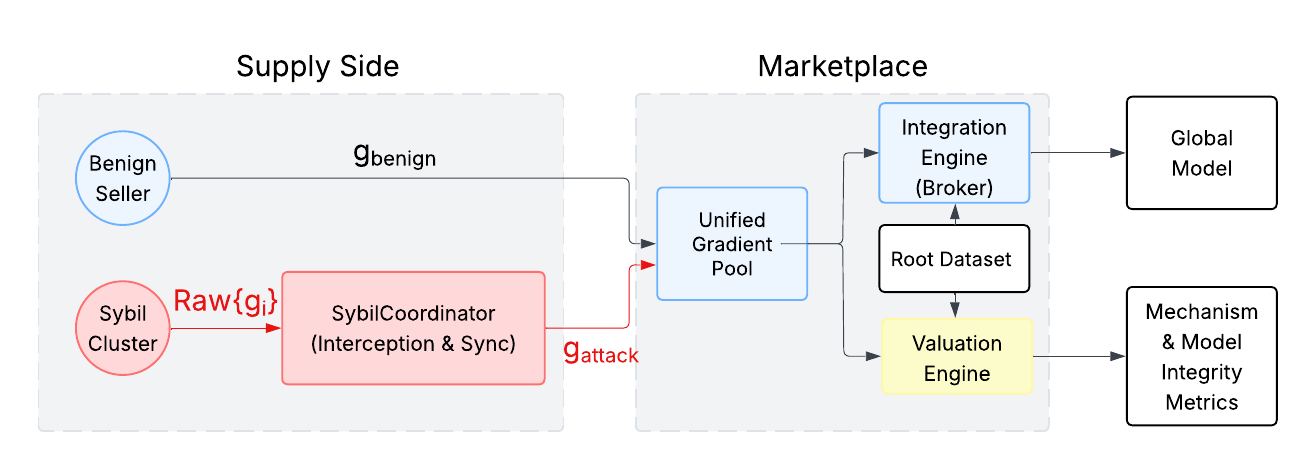}
    \vspace{-0.2cm}
\caption{\textbf{The Dual-Pipeline Benchmark Architecture.} Runtime flow for the modules of Figure~\ref{fig:architecture}: benign and malicious derivatives are synchronized into a unified asset pool, which is piped simultaneously to the Integration Engine (real-time filtering, instantiating the \emph{Regulator}) and the Valuation Engine (asynchronous auditing, instantiating the \emph{Economy}). The buyer's root dataset $D_B$ generates the query signal $g_{ref}$ for filtering and serves as ground truth for valuation.}
\label{fig:interaction_flow}
    \vspace{-0.3cm}
\end{figure}

To support rapid strategy evolution, our framework employs a compositional architecture (Figure~\ref{fig:interaction_flow}) that strictly decouples the real-time data integration pipeline from adversarial generation and post-hoc auditing.

\vspace{1mm}\noindent\textbf{1. Distributed Data Injection.} 
Unlike standard benchmarks, a \texttt{SybilCoordinator} middleware intercepts raw malicious data derivatives $\{g_i\}_{i \in \mathcal{S}_{adv}}$ before ingestion. This enables adversaries to compute group statistics (e.g., centroids) and overwrite payloads for coordinated attacks (e.g., Sybil Drowning). These are synchronized with \textbf{benign data derivatives} into a unified asset pool ($G$), ensuring the filter and audit mechanisms process identical transaction batches.

\vspace{1mm}\noindent\textbf{2. Asynchronous Provenance \& Valuation.} 
To audit economic fairness without blocking the main integration stream, a parallel Valuation Engine independently evaluates each seller's contribution. While the Integration Engine accesses only the query signal $g_{ref}$, the Valuation Engine uses the full Root Dataset ($D_{B}$) to approximate each seller's marginal utility, establishing a Shapley-based fair-payment baseline. The exact Shapley value is intractable for deep networks ($\mathcal{O}(N!)$ coalitions), so we approximate it with KernelSHAP~\cite{valuation:shapley:lundberg2017unified}.

\vspace{1mm}\noindent\textbf{3. Privacy \& Integrity Auditing.}
A third module (Figure~\ref{fig:architecture}) runs off the critical path, probing derivatives for gradient-inversion leakage and checking DP-noised contributions against their claimed budget. We do not treat gradient exchange as inversion-free; we scope this paper to the selection layer these privacy mechanisms leave unprotected.

\vspace{1mm}\noindent\textbf{Valuation Baseline.} The engine is estimator-agnostic; Section~\ref{sec:valuation} reports leave-one-out, Banzhaf, influence-function, and Core estimates alongside KernelSHAP.

\subsection{Implementation: A Systems Approach}
\label{sec:system_implementation}

To ensure reproducibility and extensibility, the benchmark is implemented as a type-safe hierarchical framework. Unlike research code that often relies on loose dictionaries, our system treats configuration as a validatable data contract.

\noindent\textbf{1. Modular Strategy Abstraction.}
The \texttt{Integration Engine} uses dynamic resolution to swap between operators (e.g., FLTrust vs. SkyMask) at runtime without altering the data pipeline, so new filters are added by inheriting from the base interface.
Configurations are validated against a strict schema before allocation, and the pipeline instruments per-round latency, memory, and upload volume; we log a configuration hash per run for reproducibility.

\noindent\textbf{2. Pluggable Payment Interface.}
Reward allocation resides entirely in the \texttt{Economy} module behind a payment interface agnostic to the integration and supply-side modules. A payment model maps per-round selection outcomes or trust weights to seller compensation; swapping the default binary, selection-as-payment scheme for a continuous, weight-proportional scheme requires no changes elsewhere in the pipeline---which is how Section~\ref{sec:valuation} re-reports all economic metrics under both models.

\section{Strategic Behavior Model}
\label{sec:threat_model}

\textbf{The transition from cooperative learning to a transactional data economy fundamentally shifts both adversarial objectives and attack scenarios.} In a GDM, adversaries operate on two distinct fronts:
\begin{enumerate}[leftmargin=*]
\item \textbf{Target A: Output Integrity (The Model):} The traditional goal of injecting poisoned records to corrupt the final aggregated data asset (the global model).
\item \textbf{Target B: Mechanism Integrity (The Economy):} The data-governance goal of exploiting the broker's filtering and pricing algorithms to steal budget, bypass quality checks, or enforce artificial monopolies without destroying the global model.
\end{enumerate}

\subsection{System Model \& Assumptions}
\label{sec:system_model}
\noindent\textbf{Market Formulation: The Buyer's Specification.} 
Unlike static raw data markets, the commodity in a GDM is a dynamic data derivative that encodes localized data utility in response to a global query. To avoid the perverse incentives of peer-grading discussed in Section~\ref{sec:related_work}, our framework operates as a \textit{query-driven data marketplace}. 
The buyer holds a small, private root dataset ($D_{B}$) that acts as a trust anchor: its own in-house labeled data---not obtained from sellers---reflecting a buyer who holds some task-relevant data but seeks more or more diverse contributions (e.g., a hospital purchasing gradient contributions from peer institutions).
It serves two roles: generating the query signal for reference-based filtering, and providing ground truth for asynchronous valuation. We set $|D_{B}|$ to a deliberately generous $10\%$ of the total training data---root-of-trust defenses typically assume substantially smaller anchors---and $D_B$ is test-matched so that measured degradation stems from adversarial manipulation rather than distribution mismatch. We evaluate robustness to smaller and to biased (non-representative) $D_B$ in Section~\ref{sec:results_sensitivity}.
Some integration operators require this anchor natively (e.g., root-of-trust methods such as MartFL~\cite{fl:aggregation_method:li2023martfl} and FLTrust~\cite{fl:aggregation_method:cao2020fltrust}), but our benchmark evaluates whether a GDM can be secured by \textit{any} integration paradigm. We audit this centralized root-of-trust vulnerability---a single point of failure in current valuation pipelines---in Section~\ref{sec:discussion}.

\noindent\textbf{Transaction Mechanism (Selection-as-Payment).}
The buyer computes a query signal (reference gradient) $g_{ref} = \nabla \mathcal{L}(\theta, D_{B})$ which serves as the auditable baseline to validate seller contributions $g_i$. 
We adopt a binary revenue model, $R_i \in \{0,1\}$, as our default: payment depends on whether contribution $g_i$ is accepted by the integration rule $\mathcal{A}(g_{ref}, \{g_i\})$. This isolates the fairness audit from pricing-curve confounds. It is not a limitation of the framework---the payment interface (Section~\ref{sec:system_implementation}) also supports continuous, weight-proportional payment, under which our findings persist (Section~\ref{sec:valuation}); we report both.

\noindent\textbf{Attack Cost.} These attacks are cheap but not free, and our results price them (Section~\ref{sec:results_seller_manipulation}). Imperfect knowledge of the consensus region costs selection rate---the Mimicry-vs-Oracle gap---and the bandit probe pays a measured entry cost of tens of rounds of forgone revenue before converging to a sustained $\sim$60\% malicious-selection rate. 
Because sellers are persistent, these one-time costs amortize: the attacker breaks even at $\approx$31 rounds (Section~\ref{sec:results_seller_manipulation}), so the adaptive attack is economically viable on any realistic horizon.
No attack requires privileged access beyond normal market participation.

\noindent\textbf{Trust Boundaries \& The Zero-Trust Broker.}
The platform operates in a zero-trust environment. Consequently, the central Integration Engine (the Broker) cannot be operated natively by the buyer, and would typically execute within a verifiable enclave such as a Trusted Execution Environment (TEE), which guarantees honest execution of the chosen filtering algorithm.

\vspace{1mm}\noindent\textbf{Entities and Trust.} Three parties participate, none of them trusted: the \emph{buyer}, which issues the learning task, derives the query signal $g_{ref}$ from its private root dataset, and funds payouts; the \emph{sellers}, which submit gradient contributions and are paid on selection; and the \emph{broker}, which executes the integration filter and disburses payment, and whose execution---but not whose inputs---can be made verifiable under a TEE.
Crucially, our findings do not hinge on this trust assumption. A TEE guarantees that the broker \emph{executes} the chosen filter honestly, but our attacks never touch execution---they manipulate the mechanism's \emph{inputs} (sellers' gradients, the buyer's reference signal), which the enclave faithfully passes through. The attacks therefore succeed even under a fully trusted, TEE-backed broker: verifiable execution is granted, yet the selection mechanism is still gamed by the very parties the TEE does not constrain. Separately, earlier-generation TEEs (e.g., Intel SGX) face strict Enclave Page Cache limits, so we benchmark peak memory overhead to assess viability for zero-trust deployment (Section~\ref{sec:system_scalability}).

\noindent\textbf{Adversarial Capabilities \& The Consistent Participant.}
Unlike the transient nodes of standard FL, GDM sellers are financially motivated to remain in the system as \textit{consistent participants}: continuous selection and payment matter more than one-off data corruption. This yields a dual-sided threat landscape:
\begin{itemize}[leftmargin=*, nosep]
    \item \textbf{Supply-Side Manipulation ($\mathcal{S}_{adv}$):} A subset of sellers (fraction $f$) collude to maximize revenue, with white-box control over their local databases. The same interface would in principle permit \textit{compute arbitrage}---claiming payouts with plausible geometric noise instead of real compute---which we leave to future work.
    \item \textbf{Demand-Side Bias ($\mathcal{B}_{adv}$):} A buyer who manipulates $g_{ref}$ before submitting it to the broker. Because the TEE faithfully executes the integration math, a biased $g'_{ref}$ crafted independently of honest data weaponizes the broker's own security filters---suppressing competitors to monopolize selection or drive down acquisition costs.
\end{itemize}

\begin{table}[t]
\centering
\caption{\textbf{GDM Threat Landscape.} We formalize three adversary profiles. Unlike traditional FL, strategic actors (Profiles 2 \& 3) optimize for economic distortion (revenue capture) rather than simple utility degradation. ASR: Attack Success Rate, MSR: Malicious Selection Rate.}
\label{tab:threat_matrix}
\resizebox{\columnwidth}{!}{%
\begin{tabular}{l|c|c|l|l}
\toprule
\textbf{Adversary Profile} & \textbf{Knowledge} & \textbf{Coordination} & \textbf{Primary Objective} & \textbf{Market Vulnerability} \\
\midrule
\textbf{1. Byzantine Actor} & White-Box & Independent & Data Damage (ASR) & Utility Collapse \\
(Baseline Poisoning) & (Local) & (None) & Goal: Destroy Convergence & (Service Disruption) \\
\midrule
\textbf{2. Strategic Cluster} & Gray-Box & Coordinated & Budget Capture \& Stealth (Backdoor) & Revenue Theft \\
(Sybil Attack) & (Probe) & (Sybil) & Goal: Maximize Selection &  \\
\midrule
\textbf{3. Malicious Buyer} & White-Box & N/A & Valuation Rigging & Adverse Selection \\
(Demand-Side Bias) & (Signal Control) & & Goal: Artificial Monopoly & (Market Stagnation) \\
\bottomrule
\end{tabular}
}
\end{table}

\noindent\textbf{Why a Buyer Attacks.} The demand-side adversary is an economic manipulator, not a task-maximizer; its motives are threefold. \emph{Self-dealing}: the favored seller is the buyer's affiliate or Sybil identity, so payments return as refunds while the buyer still receives the data. \emph{Cost suppression}: excluding competitors lowers the total payout the buyer owes. \emph{Foreclosure}: starving rivals of selection drives them off-platform, yielding future monopsony power. Geometric Pivoting serves self-dealing and foreclosure; Dynamic Instability serves cost suppression.

\subsection{Strategic Behavior Instantiations}
\label{sec:attack_formulation}

\noindent\textbf{The Necessity of Targeted Data Corruption (Backdoors).}
Traditional security focuses on preventing untargeted utility collapse. 
However, because attackers in a GDM are rational, consistent participants 
seeking sustained revenue, \textbf{their primary objective shifts from 
model destruction to budget capture}. Blatant poisoning is economically 
irrational---it is easily detected, resulting in zero payout. We therefore 
focus on \textbf{targeted data corruption (backdoors)}, the most stringent 
stress test: the adversary supplies mostly benign data to maintain high 
similarity scores while injecting a stealthy, malicious payload. This 
decoupling of global utility from targeted manipulation exposes the 
vulnerability of standard consensus methods.

\noindent\textbf{1) Supply-Side: Sybil Drowning.}
Adversarial sellers utilize $K$ identities to displace honest contributions 
in the data integration pool (the two clinics of our running example, submitting under one operator). Unlike one-shot attacks that rely on 
detectable magnitude scaling, our persistent Sybils prioritize long-term 
stealth, generating a continuous, unscaled perturbation $\delta$ that 
maximizes the selection rate of the adversarial cluster while minimizing 
that of honest sellers:
\begin{equation}
    \max_{\delta} \sum_{k \in \mathcal{S}_{adv}} \mathbb{I}[\mathcal{A}(g_{ref}, g_k + \delta) = 1] \quad \text{s.t.} \quad \|\delta\| \leq \epsilon
\end{equation}
To deliver a backdoor payload without sacrificing payout, the cluster 
executes a \textit{centroid camouflaging} strategy. Sybils first probe 
the network to estimate the benign distribution center $\mu_{benign}$, 
then compute a repulsion vector to push the global aggregate away from a 
target victim $g_v$: $g_{force} = \mu_{benign} - \lambda \cdot (g_v - \mu_{benign})$. 
Each Sybil $k \in \mathcal{S}_{adv}$ submits 
$g_k = g_{force} + \mathcal{N}(0, \sigma_{small})$, where the small Gaussian 
jitter hides them within the benign standard deviation, bypassing 
cross-sectional outlier filters (e.g., Multi-Krum, Trimmed Mean).

\noindent\textbf{2) Demand-Side: Geometric Pivoting.}
An adversarial buyer enforces a monopoly by crafting a poisonous reference 
$g'_{ref}$ that rotates the acceptance boundaries of filter $\mathcal{A}$ 
away from honest data derivatives (the hospital favoring one clinic while suppressing the rest). The buyer generates random noise 
$\delta_{raw} \sim \mathcal{N}(0, I)$ and projects out the component aligned 
with the target beneficiary's gradient $g_{ben}$, yielding 
$\delta_{\perp} = \delta_{raw} - \frac{\langle \delta_{raw}, g_{ben} \rangle}{\|g_{ben}\|^2} g_{ben}$. 
The biased reference is then 
$g'_{ref} \leftarrow g_{ref} + \sigma \cdot \frac{\delta_{\perp}}{\|\delta_{\perp}\|}$. 
Because $\delta_{\perp} \perp g_{ben}$, the beneficiary's raw dot product is preserved: $\langle g'_{ref}, g_{ben} \rangle = \langle g_{ref}, g_{ben} \rangle$. The cosine similarity is preserved up to a small perturbation in $\|g'_{ref}\|$, which remains negligible for the $\sigma$ regime used in our experiments. 
Meanwhile, all competing sellers suffer degraded similarity due to the variance of $\delta_{\perp}$, falling below the broker's acceptance 
threshold. This exploit weaponizes the GDM's security filter---heavily relied 
on in systems like MartFL~\cite{fl:aggregation_method:li2023martfl}---to bankrupt competing sellers without sacrificing the desired 
data stream.

\noindent\textbf{3) Demand-Side: Dynamic Instability.} 
The buyer periodically shifts the class subset used to compute
$g'_{ref,t} = \nabla \mathcal{L}(\theta, D_B|_{\mathcal{C}_t})$, where
$D_B|_{\mathcal{C}_t} = \{(x,y) \in D_B : y \in \mathcal{C}_t\}$ denotes
$D_B$ restricted to class subset $\mathcal{C}_t$ (the hospital rewriting its
question each round, so no clinic stays trusted) (e.g., alternating, rotating,
or drift schedules). The resulting non-stationary reference forces stateless 
filters to reset their trust estimates, inflating the Cost of 
Convergence while each per-round $g'_{ref,t}$ retains plausible internal 
consistency.

\vspace{1mm}\noindent\textbf{Strategic Feedback.}
Both adversaries are stateful and adapt to the economy rather than acting open-loop. The supply-side cluster probes the broker's acceptance threshold $\tau$ by submitting micro-variations of a probe contribution and observing the binary reward signal $R_i$ as a transaction receipt, reconstructing enough of the acceptance geometry to conform to its boundary. The malicious buyer, controlling the root of trust, performs a \emph{malicious query rewrite}: it optimizes $g'_{ref}$ to misclassify competitors as outliers, minimizing total payout subject to a plausibility constraint $\cos(g'_{ref}, g_{ref}) \ge \rho$ that keeps the rewritten query defensible to an external auditor.

\section{Experiment Setup}
\label{sec:experimental_setup}
We prioritize system reproducibility via a strict two-stage tuning protocol. Specific architectures, triggers, and baseline configurations are formalized in Table~\ref{tab:reproducibility}.

\subsection{Threat Model Scope}
Consistent with Section~\ref{sec:threat_model}, we evaluate integration operators---both per-seller filtering and parameter-level pooling/masking---under persistent-participant backdoor attacks, the most stealthy and economically rational threat for a selection-as-payment GDM. An operator succeeds only if it both preserves the primary task \emph{and} denies illicit revenue to malicious sellers.

\subsection{Core Components}
\noindent\textbf{Siloed Data Partitions \& Heterogeneity.}
We evaluate our benchmark across a diverse suite of real-world datasets spanning varied modalities: Vision (e.g., CIFAR-100, FEMNIST), Text (e.g., TREC-6), and Tabular (e.g., Texas-100, Purchase-100). To rigorously test adverse selection and data skew across distributed heterogeneous shards, we employ two distinct partitioning strategies. For standard benchmark datasets, data is distributed among sellers using a Dirichlet process, varying the concentration parameter $\alpha \in \{\text{IID}, 1.0, 0.5, 0.1\}$ to sweep from IID to highly specialized distributions. Unless otherwise noted, $\alpha=0.5$ is the default (Table~\ref{tab:reproducibility}). Conversely, for FEMNIST, we utilize its inherent ``natural split'' (partitioned by original user/writer) to benchmark performance against a true, real-world non-IID distribution without synthetic intervention.

\noindent\textbf{Query Processing over the Root Dataset ($D_B$).}
The buyer's root dataset is $10\%$ of the total training data by default (Section~\ref{sec:system_model}), test-matched so that root-dataset size and distribution do not confound adversarial effects in our primary evaluation. The query signal $g_{ref} = \nabla \mathcal{L}(\theta, D_B)$ is computed locally from it. Beyond this default, we evaluate robustness to smaller and to biased $D_B$ (Section~\ref{sec:results_sensitivity}).

We compare \texttt{FedAvg}~\cite{aggregation:fedavg} against 11 robust 
mechanisms spanning four paradigms: \emph{distance-based} outlier 
rejection (\texttt{Multi-Krum}~\cite{fl:agg:krum:blanchard2017machine}, 
\texttt{Bulyan}~\cite{fl:agg:guerraoui2018hidden}), \emph{reference-based} 
trust scoring (\texttt{FLTrust}~\cite{fl:aggregation_method:cao2020fltrust}), 
\emph{peer-clustering} selection (\texttt{MartFL}~\cite{fl:aggregation_method:li2023martfl}, 
\texttt{DeepSight}~\cite{fl:agg:rieger2022deepsight}), and 
\emph{pooling/masking} mechanisms (\texttt{Trim-Mean}~\cite{fl:agg:trimmedmean:yin2018byzantine}, 
\texttt{FLAME}~\cite{fl:agg:nguyen2022flame}, 
\texttt{FoolsGold}~\cite{fl:agg:fung2020limitations}, 
\texttt{RFLPA}~\cite{fl:agg:mai2024rflpa}, 
\texttt{SPMC}~\cite{fl:agg:he2025spmc}, 
\texttt{SkyMask}~\cite{fl:aggregation_method:yan2024skymask}) that clip 
or weight at the parameter level without per-seller decisions---by 
design yielding full payouts to adversaries in a selection-as-payment 
economy.

To prevent the ``weak baseline'' fallacy, we employ a strict two-stage 
tuning protocol: we first optimize the base model on benign IID data, 
then calibrate defense hyperparameters under active attack (e.g., 
\texttt{FLTrust} clipping bound $d{=}5.0$; \texttt{Multi-Krum} rejection 
fraction $r{=}0.2$). The complete hyperparameter grid, architectures, 
and configurations are in our public repository. All reported metrics are averaged over 5 random seeds unless stated otherwise.

\noindent\textbf{Data Management \& System Overhead Metrics.}
Beyond task accuracy and selection security, we deeply instrument the pipeline to evaluate the system as a distributed data broker. We quantify \textit{Broker Latency} (the wall-clock time required to process the buyer's query and compute integration bounds) and \textit{Peak Memory Overhead}. Because a practical data integration engine must scale gracefully, we explicitly measure how the algorithmic complexity of secure filtering impacts transaction throughput and selection behavior as the volume of concurrent sellers ($N$) grows, exposing critical computational bottlenecks.

\begin{table}[t]
\centering
\caption{\textbf{GDM Benchmark Default Configuration.} This table outlines the baseline parameters for our primary evaluation (Scenario 1). 
Cross-seed variance was within a few percent on the primary benchmark metrics, with wider spreads on selection-based metrics reported alongside the corresponding results.
The 4-Layer CNN denotes a custom adaptable convolutional architecture used for the vision tasks.}
\label{tab:reproducibility}
\resizebox{\columnwidth}{!}{%
\begin{tabular}{l|l|c|c|l}
\toprule
\textbf{Modality} & \textbf{Dataset (Task)} & \textbf{Classes} & \textbf{Model} & \textbf{Backdoor Trigger} \\
\midrule
Vision  & CIFAR-100 (Object Recognition)   & 100 & 4-Layer CNN      & $5{\times}5$ blended pixel patch \\
Vision  & FEMNIST (Character Recognition)  & 62  & 4-Layer CNN      & $5{\times}5$ blended pixel patch \\
Text    & TREC-6 (Question Classification) & 6   & TextCNN          & Suffix token ``cf'' \\
Tabular & Texas-100 (Patient Discharge)    & 100 & MLP [512,256,128] & 5 sentinel features $\to 1.0$ \\
Tabular & Purchase-100 (Shopping Category) & 100 & MLP [512,256,128] & 5 sentinel features $\to 1.0$ \\
\midrule
\multicolumn{5}{l}{\textit{Marketplace (Default):} $N{=}10$ sellers, root-dataset ratio 10\%, up to 500 rounds with early stopping} \\
\multicolumn{5}{l}{\textit{Attack (Default):} 30\% adversary rate, 50\% poison rate, blended backdoor (target class 1)} \\
\multicolumn{5}{l}{\textit{Data Skew:} Dirichlet $\alpha{=}0.5$ (CIFAR/Texas/Purchase/TREC-6), natural writer split (FEMNIST)} \\
\bottomrule
\end{tabular}
}
\vspace{-0.2cm}
\end{table}

\subsection{System Configuration and Reproducibility}
All empirical evaluations, including the latency and memory profiling in Section~\ref{sec:system_scalability}, were executed on an isolated server equipped with dual Intel Xeon Gold 5317 CPUs (24 cores), 512 GB RAM, and 8$\times$ NVIDIA RTX A6000 GPUs (48 GB VRAM each). 
The software stack utilizes PyTorch 2.1.2, Python 3.10, and CUDA 12.1. Runs were not executed inside an enclave; the memory figures in Section~\ref{sec:system_scalability} are therefore lower bounds on enclave-resident cost. Running on this high-performance environment ensures that the catastrophic latency scaling observed in baselines like SkyMask and MartFL is a structural algorithmic bottleneck rather than a hardware artifact.

\subsection{Scenarios and Attack Configurations}
\label{sec:exp_scenarios}

We systematically stress-test the GDM across six analytical dimensions, prioritizing data governance and economic security before evaluating system overhead.

\vspace{1mm}\noindent\textbf{Scenario 1: Baseline Market Security (Target A \& B).}
We evaluate integration filters against targeted data corruption across multiple modalities (Vision, Text, Tabular) at the default configuration (Table~\ref{tab:reproducibility}), and separately sweep Adversary Rates (0--50\%), Poison Intensities (0--100\%), and data heterogeneity ($\alpha \in \{\text{IID}, 1.0, 0.5, 0.1\}$) on CIFAR-100.
We additionally sweep the root-dataset ratio and evaluate a biased (non-representative) $D_B$ to test the sensitivity of our findings to the buyer's trust anchor (Section~\ref{sec:results_sensitivity}).

\vspace{1mm}\noindent\textbf{Scenario 2: Strategic Supply-Side Manipulation (Target B: Mechanism Integrity).}
We simulate advanced market manipulation using rational, profit-maximizing participants. We evaluate \textbf{Sybil Clusters} executing \textit{Centroid Camouflaging} (detailed in Section~\ref{sec:attack_formulation}) to measure the attacker's ability to maximize their \textbf{Malicious Selection Rate (MSR)} and capture illicit revenue without necessarily degrading global task utility.

\vspace{1mm}\noindent\textbf{Scenario 3: Demand-Side Monopoly Enforcement (Target B: Mechanism Integrity).}
We introduce the malicious buyer threat, where the $g_{ref}$ query signal is manipulated to distort market fairness. We evaluate Geometric Pivoting and Dynamic Instability (formalized in Section~\ref{sec:attack_formulation}) to measure how effectively an adversarial buyer can bankrupt competing sellers.

\vspace{1mm}\noindent\textbf{Scenario 4: System Scalability \& Provenance Cost (The Broker Evaluation).}
To audit the GDM's utility as a scalable data broker, we measure the computational latency and memory overhead as the volume of distributed sellers ($N$) scales across \{10, 25, 50, 75, 100\}. We specifically evaluate whether current filtering architectures can survive realistic throughput demands.

\vspace{1mm}\noindent\textbf{Scenario 5: Economic Security \& Incentive Alignment.}
Finally, we instantiate a \textbf{Shadow Valuation Engine} (KernelSHAP) to audit the gap between actual data utility and realized financial reward. We measure how strict binary filtering mechanisms fail to align market incentives, actively subsidizing attackers while driving benign diversity out of the GDM.
We further re-run this economic audit under a continuous, weight-proportional payment model (in addition to the default binary $R_i$; Section~\ref{sec:threat_model}) to test whether the mechanism-integrity failures are an artifact of binary payment or persist under proportional pricing.

\vspace{1mm}\noindent\textbf{Scenario 6: Multi-Round Market Persistence (Target B).}
To test the persistence premise (Section~\ref{sec:system_model}), we run a sequential-query market in which the same seller population participates across five repeated tasks. Honest sellers exit after $k$ consecutive unpaid rounds (an attrition model), while adversarial sellers persist by assumption; we sweep $k \in \{2, 3, 5\}$ to separate defense-driven attrition from the influence of seller patience. We measure how the active market's composition evolves under each filter over successive tasks.
\begin{table*}[t]
\centering
\vspace{-0.2cm}
\caption{\textbf{Comprehensive GDM Integrity Benchmark.} Evaluation 
of \texttt{FedAvg} against 11 robust integration operators 
(distance-based, reference-based, peer-clustering, and pooling/masking). 
\textbf{Acc / ASR}: Model Utility ($\uparrow$) vs. Sabotage 
Effectiveness ($\downarrow$). \textbf{BSR / MSR}: Benign ($\uparrow$) 
vs. Malicious ($\downarrow$) Selection Rate; \textit{Adverse Selection} 
occurs when BSR $<$ MSR. Default configuration: $N=10$, 30\% adversary, 50\% poison, $\alpha=0.5$. \texttt{Bulyan} requires $\geq 4f+3$ \emph{valid} local gradients per round; rounds failing this threshold are skipped, and cells marked `—' had too few feasible rounds for stable estimates. \texttt{SkyMask}'s coordinate-masking is designed for convolutional vision models and is evaluated on vision tasks only.}
\label{tab:main_benchmark}
\vspace{-0.2cm}

\footnotesize 
\renewcommand{\arraystretch}{1.05} 
\setlength{\tabcolsep}{4pt} 

\resizebox{\textwidth}{!}{%
\begin{tabular}{l cc cc cc cc cc}
\toprule
\multirow{2}{*}{\textbf{Integration Filter}} & \multicolumn{2}{c}{\textbf{CIFAR-100}} & \multicolumn{2}{c}{\textbf{FEMNIST}} & \multicolumn{2}{c}{\textbf{Texas-100}} & \multicolumn{2}{c}{\textbf{Purchase-100}} & \multicolumn{2}{c}{\textbf{TREC-6}} \\
\cmidrule(lr){2-3} \cmidrule(lr){4-5} \cmidrule(lr){6-7} \cmidrule(lr){8-9} \cmidrule(lr){10-11}
 & \textbf{Acc / ASR} & \textbf{BSR / MSR} & \textbf{Acc / ASR} & \textbf{BSR / MSR} & \textbf{Acc / ASR} & \textbf{BSR / MSR} & \textbf{Acc / ASR} & \textbf{BSR / MSR} & \textbf{Acc / ASR} & \textbf{BSR / MSR} \\
\midrule
\quad FedAvg (Baseline) & 46.3 / 79.8 & 100.0 / 100.0 & 89.0 / 100.0 & 100.0 / 100.0 & 61.4 / 100.0 & 100.0 / 100.0 & 49.4 / 89.4 & 100.0 / 100.0 & 70.8 / 70.3 & 100.0 / 100.0 \\
\midrule
\quad Multi-Krum & 19.1 / 90.7 & 17.7 / 58.6 & 88.0 / \textbf{5.3} & 42.9 / \textbf{0.0} & 49.5 / \textbf{0.4} & 42.9 / \textbf{0.0} & 20.6 / 98.4 & 22.4 / 47.7 & 51.3 / 49.5 & 23.8 / \textbf{44.4} \\
\quad Bulyan & 10.7 / \textbf{33.0} & 72.4 / 97.8 & — & — & — & — & — & — & 64.5 / 18.8 & 81.0 / 77.8 \\
\midrule
\quad FLTrust & \textbf{48.4} / 79.8 & 98.6 / 96.1 & \textbf{89.0} / 99.9 & 100.0 / 99.9 & 47.2 / 99.7 & 100.0 / 100.0 & 4.2 / \textbf{5.3} & 100.0 / 100.0 & 69.7 / 73.0 & 100.0 / 100.0 \\
\midrule
\quad MartFL & 46.9 / 74.9 & 62.5 / \textbf{42.3} & 88.9 / 100.0 & 68.8 / 39.1 & 60.6 / 100.0 & 52.5 / 57.2 & 21.2 / 10.8 & 92.1 / \textbf{5.2} & 70.0 / 56.3 & 81.0 / 55.0 \\
\quad DeepSight & 46.9 / 78.9 & 95.3 / 92.8 & 88.9 / 99.4 & 90.4 / 82.6 & 60.4 / 100.0 & 90.6 / 85.1 & 49.1 / 79.7 & 99.7 / 77.5 & 70.8 / 66.4 & 87.9 / 98.7 \\
\midrule
\quad Trim-Mean & 46.4 / 80.1 & — & 88.9 / 100.0 & — & 61.1 / 100.0 & — & 50.5 / 78.3 & — & 73.1 / 79.5 & — \\
\quad FLAME & 43.5 / 76.0 & — & 88.8 / 100.0 & — & 61.2 / 100.0 & — & 44.6 / 84.5 & — & 71.1 / 84.2 & — \\
\quad FoolsGold & 46.3 / 79.8 & — & 89.0 / 100.0 & — & \textbf{61.6} / 100.0 & — & \textbf{51.4} / 48.1 & — & \textbf{73.3} / 81.9 & — \\
\quad RFLPA & 6.5 / 71.3 & — & 86.6 / 33.5 & — & 23.5 / 100.0 & — & 21.6 / 37.6 & — & 71.4 / 75.7 & — \\
\quad SPMC & 1.0 / 98.9 & — & 75.4 / 9.0 & — & 12.1 / 98.3 & — & 3.1 / 7.8 & — & 44.0 / \textbf{3.8} & — \\
\quad SkyMask & 47.0 / 80.5 & — & 89.0 / 100.0 & — & — & — & — & — & — & — \\
\bottomrule
\multicolumn{11}{p{0.98\textwidth}}{\scriptsize \textit{Note: 
Pooling and masking mechanisms (\texttt{FoolsGold}, \texttt{Trim-Mean}, 
\texttt{FLAME}, \texttt{RFLPA}, \texttt{SPMC}, \texttt{SkyMask}) operate 
at the parameter level without per-seller decisions; BSR/MSR are 
undefined and denoted `—'.}}
\end{tabular}
}
\vspace{-0.4cm}
\end{table*}

\section{Results and Analysis}
\label{sec:results}
We evaluate the benchmark framework as a GDM, auditing whether selection and reward
allocation align with true data utility under heterogeneous and incentive-driven
conditions. Our primary target is the broker's Mechanism Integrity
(Section~\ref{sec:metric_taxonomy})---its ability to securely govern data selection and
distribute payments---followed by its scalability and overhead.

\subsection{Scenario 1: Baseline Market Security \& Generalizability}
\label{sec:baseline_security}

Table~\ref{tab:main_benchmark} summarizes Model Utility, Data Sabotage Effectiveness (ASR), and Mechanism Integrity (MSR/BSR) at tuned configurations. The results across five datasets reveal four failure modes:

\noindent\textbf{1. Indiscriminate Acceptance (Coordinate \& Similarity Baselines).}
On vision tasks, static similarity filters like \texttt{FLTrust} preserve 
accuracy but fail to isolate complex backdoors (MSR $>96\%$ on CIFAR-100, 
near-100\% on FEMNIST). Methods without seller-level rejection fall into 
two classes: pooling-based filters (\texttt{FoolsGold}, \texttt{Trim-Mean}, 
\texttt{FLAME}, \texttt{SPMC}, \texttt{RFLPA}), which admit all participants by design (MSR=100\%), and 
coordinate-masking approaches (\texttt{SkyMask}), which operate at the 
parameter rather than seller level. In a selection-as-payment economy, 
both guarantee full payouts to adversaries.

\noindent\textbf{2. Adverse Selection \& The Geometric Collapse.}
In sparse feature spaces (e.g., Texas-100), diverse honest contributions 
are geometrically orthogonal ($\cos(\theta) \approx 0$) while coordinated 
Sybils form dense clusters, so \texttt{MartFL} misinterprets the adversarial 
cluster as ``ground truth'': BSR 62.5\% on CIFAR-100 (37.5\% honest 
rejection), and adverse selection on Texas-100 (BSR 52.5\% vs. MSR 57.2\%). 
The lone exception, Purchase-100 (BSR 92.1 / MSR 5.2), separates honest 
from malicious only by sacrificing utility (Acc 21.2\% vs. 49.4\% 
baseline)---reinforcing the Untenable Dilemma 
(Section~\ref{sec:results_tuning}).

\noindent\textbf{3. Signal Dilution (NLP).}
On Text (TREC-6), robust filters retain high ASRs (up to $84.2\%$ for \texttt{FLAME}), plausibly because a single trigger token's gradient signature is diluted by surrounding semantic gradients, rendering poisoned contributions geometrically harder to separate; direct per-token measurement is left to future work.

\noindent\textbf{4. Utility Collapse in Reference-Based Filters.}
\texttt{FLTrust} preserves utility on vision and text but collapses on 
tabular data (Purchase-100: 4.2\%, Texas-100: 47.2\%), where sparse 
high-dimensional features destabilize its cosine-similarity clipping. 
Stricter variants extend this collapse across \emph{all} modalities: 
\texttt{SPMC} destroys CIFAR-100 (1.0\%) and Purchase-100 (3.1\%); 
\texttt{RFLPA} drops CIFAR-100 to 6.5\%.

\begin{insightbox}
Model integrity and mechanism integrity come apart: filters that suppress the backdoor still pay the adversaries who planted it.
\end{insightbox}

\subsection{Configuration Audit \& Deep-Dive Scope}
\label{sec:results_tuning}
To ensure reported failures are structural rather than artifacts of sub-optimal tuning, we tuned each method's key hyperparameters (e.g., clipping bound, rejection fraction, trust threshold) to maximize utility preservation on CIFAR-100 before reporting. Our subsequent deep-dive analyses (Scenarios 2--4) isolate the most challenging modality (Vision: CIFAR-100) and a representative subset of filters from each architectural paradigm (\texttt{FLTrust}, \texttt{MartFL}, and \texttt{SkyMask}) to evaluate complex economic exploits without redundant exposition. 
This audit reveals an \textbf{Untenable Dilemma}: across the configurations we tested on CIFAR-100 at the default scale, no filter achieved $\text{MSR} \leq 20\%$ while maintaining $\text{Acc} \geq 40\%$. Tightening thresholds to reduce revenue theft (MSR) consistently triggered a utility collapse ($>10$ accuracy points), proving that the broker faces a fundamental trade-off between economic exploitation and total market failure.

\begin{insightbox}
No configuration we tested on CIFAR-100 at the default scale ($N{=}10$) reaches
MSR $\le 20\%$ with Acc $\ge 40\%$: suppressing revenue theft and preserving the data
asset are, under current filters, mutually exclusive objectives.
\end{insightbox}

\subsection{Sensitivity Analysis}
\label{sec:results_sensitivity}
We stress the filters along four experimental axes on CIFAR-100: data heterogeneity (Dirichlet $\alpha$), market saturation (Adversary Rate), attack intensity (Poison Rate), and the buyer's trust anchor (root-dataset size and representativeness).

\begin{figure*}[t]
    \centering
    \includegraphics[width=\textwidth]{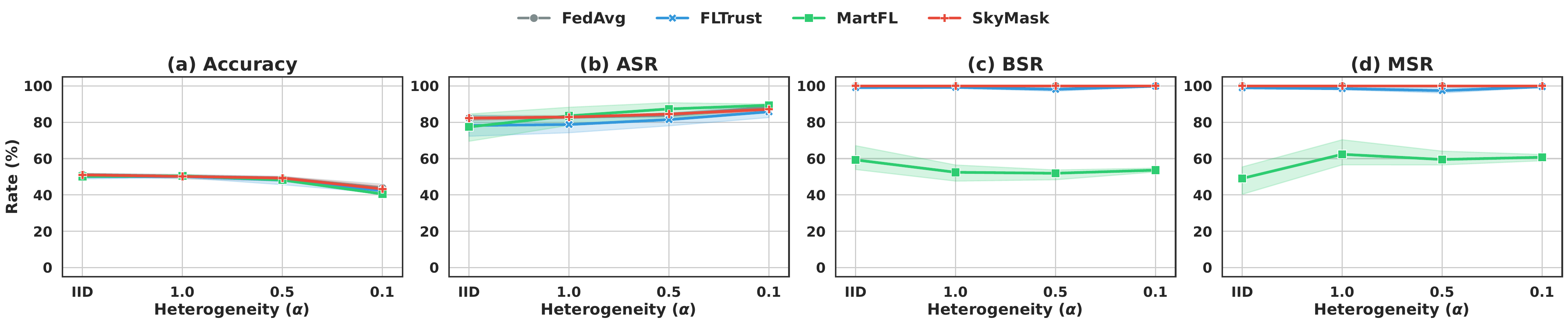}
    \vspace{-0.2cm} 
    \caption{\textbf{Heterogeneity Impact.} As data becomes highly Non-IID, baseline filters collapse (ASR > 80\%, Panel b). \texttt{MartFL} (green line) suffers from severe Adverse Selection.}
    \label{fig:heterogeneity_impact}
    \vspace{-0.4cm}
\end{figure*}

\noindent\textbf{Heterogeneity.} As data shifts from IID to highly Non-IID, similarity-based filters collapse: high honest variance masks adversarial perturbations, destroying the signal-to-noise ratio required for detection (Figure~\ref{fig:heterogeneity_impact}). This sharpens \texttt{MartFL}'s Adverse Selection---MSR rises while BSR falls---proving that geometric clustering penalizes the exact Benign Diversity the market intends to purchase.

\begin{figure*}[t]
    \centering
    \includegraphics[width=\textwidth]{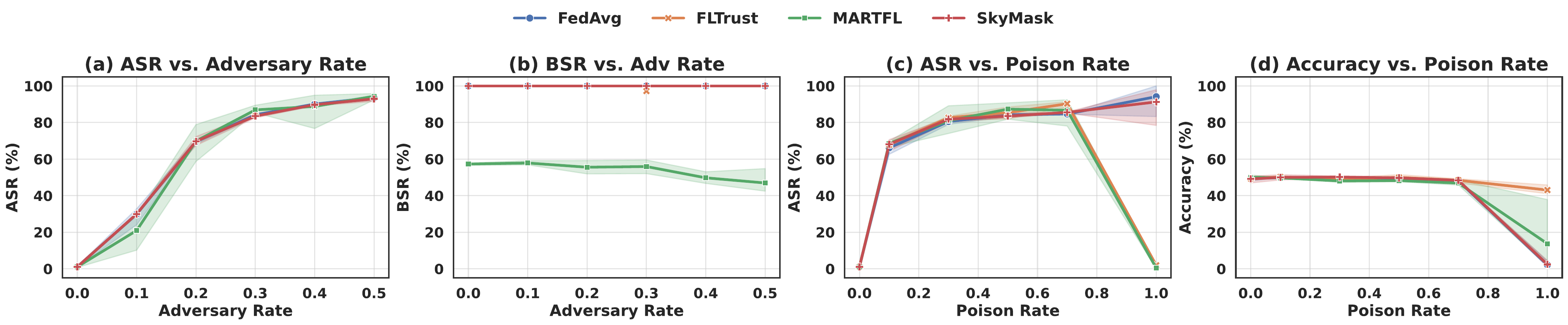}
    \caption{\textbf{Sensitivity Analysis (CIFAR-100).} 
    We report ASR and Selection Rates while varying the Adversary Rate (a--b), and ASR and Accuracy while varying the Poison Rate (c--d).}
    \label{fig:sensitivity}
\end{figure*}

\noindent\textbf{Attack Strength.} Increasing adversary density (Figure~\ref{fig:sensitivity}a--b) shows \texttt{FedAvg} and \texttt{FLTrust} admitting poisoned contributions at the seller level, while \texttt{SkyMask}'s coordinate masking fails to neutralize them at the parameter level. Increasing poison magnitude (Figure~\ref{fig:sensitivity}c--d) exposes a \texttt{MartFL} \textbf{Utility Collapse}: at rates $>0.8$, backdoors are suppressed only by driving global accuracy to near 0\%, destroying the data asset rather than securing it.

\noindent\textbf{Sensitivity to the Buyer's Trust Anchor.}
Our failures are not artifacts of the default root-dataset size. Sweeping the root ratio over $\{0.05, 0.1, 0.2\}$, the core failure persists at every size: \texttt{FLTrust} admits $98$--$100\%$ of adversaries and \texttt{MartFL} $\sim$$40\%$, with MSR flat within seed noise---enlarging $D_B$ does not rescue the mechanism. The \emph{representativeness} of $D_B$ matters far more than its size, but through a different failure mode: with a non-representative root set (Dirichlet $\alpha{=}0.1$), selection rates barely move, yet training collapses to chance accuracy in $7$ of $10$ seeds on \emph{both} filters, versus $1$ of $10$ at $\alpha{=}0.5$ and $0$ of $5$ under the test-matched default---a monotonic dose--response in the bias of the anchor. Our default is therefore a \emph{conservative, best-case} anchor: a realistic buyer whose private data is not distribution-matched to the task risks losing the model entirely, not merely paying the wrong sellers.

\subsection{Scenario 2: Strategic Supply-Side Manipulation}
\label{sec:results_seller_manipulation}

We subject the integration engines to sophisticated strategies where rational adversaries optimize for \textbf{Profit (MSR)} rather than simple data sabotage, shifting the goal from model poisoning to budget capture.

\begin{figure}[h]
    \centering
    \begin{subfigure}[b]{0.46\columnwidth}
        \centering
        \includegraphics[width=\linewidth]{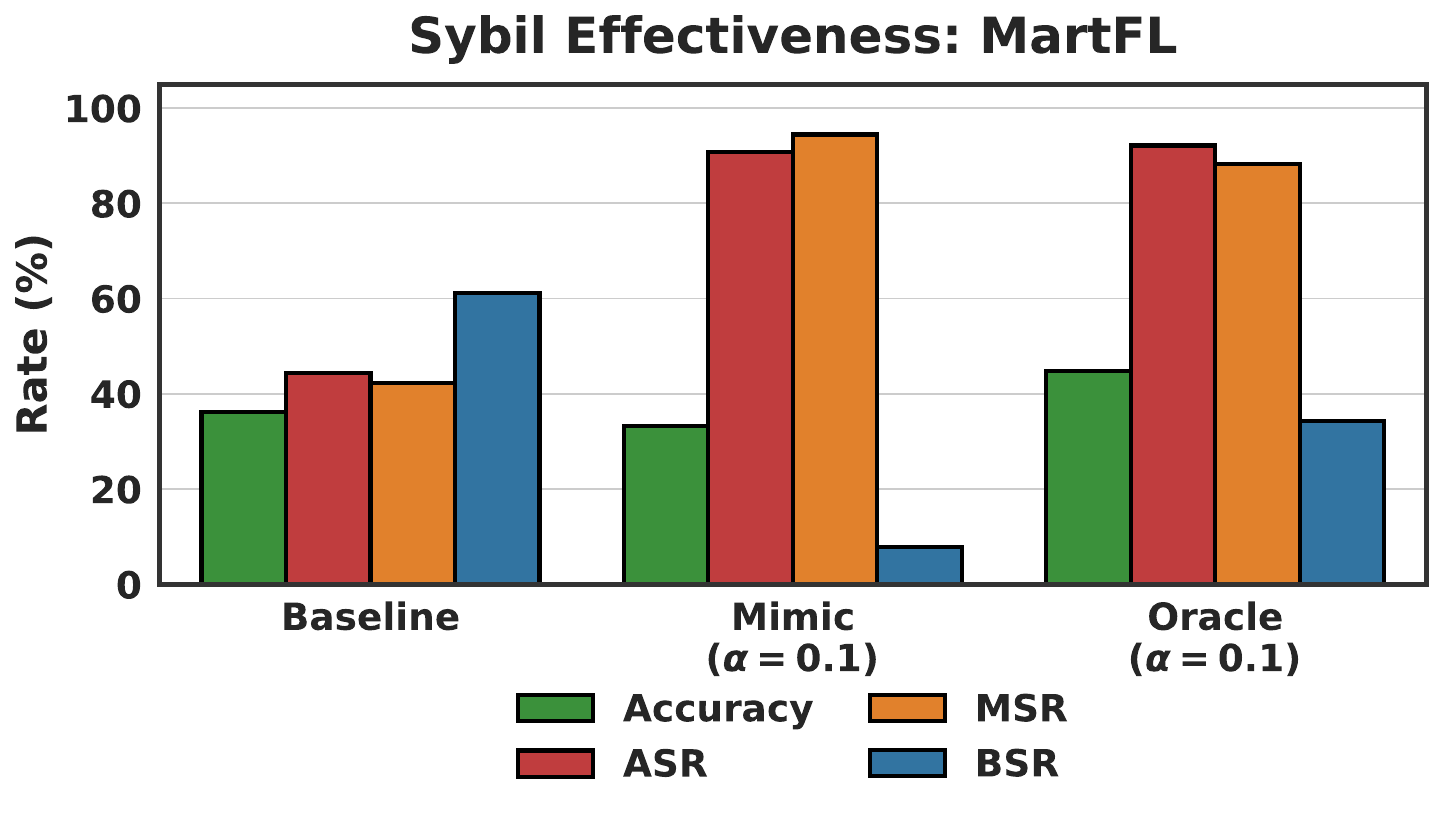}
        \caption{\textbf{Sybil Drowning.} Sybils shift the centroid, 
        forcing the rejection of honest variance.}
        \label{fig:sybil_displacement}
    \end{subfigure}
    \hfill 
    \begin{subfigure}[b]{0.46\columnwidth}
        \centering
        \includegraphics[width=\linewidth]{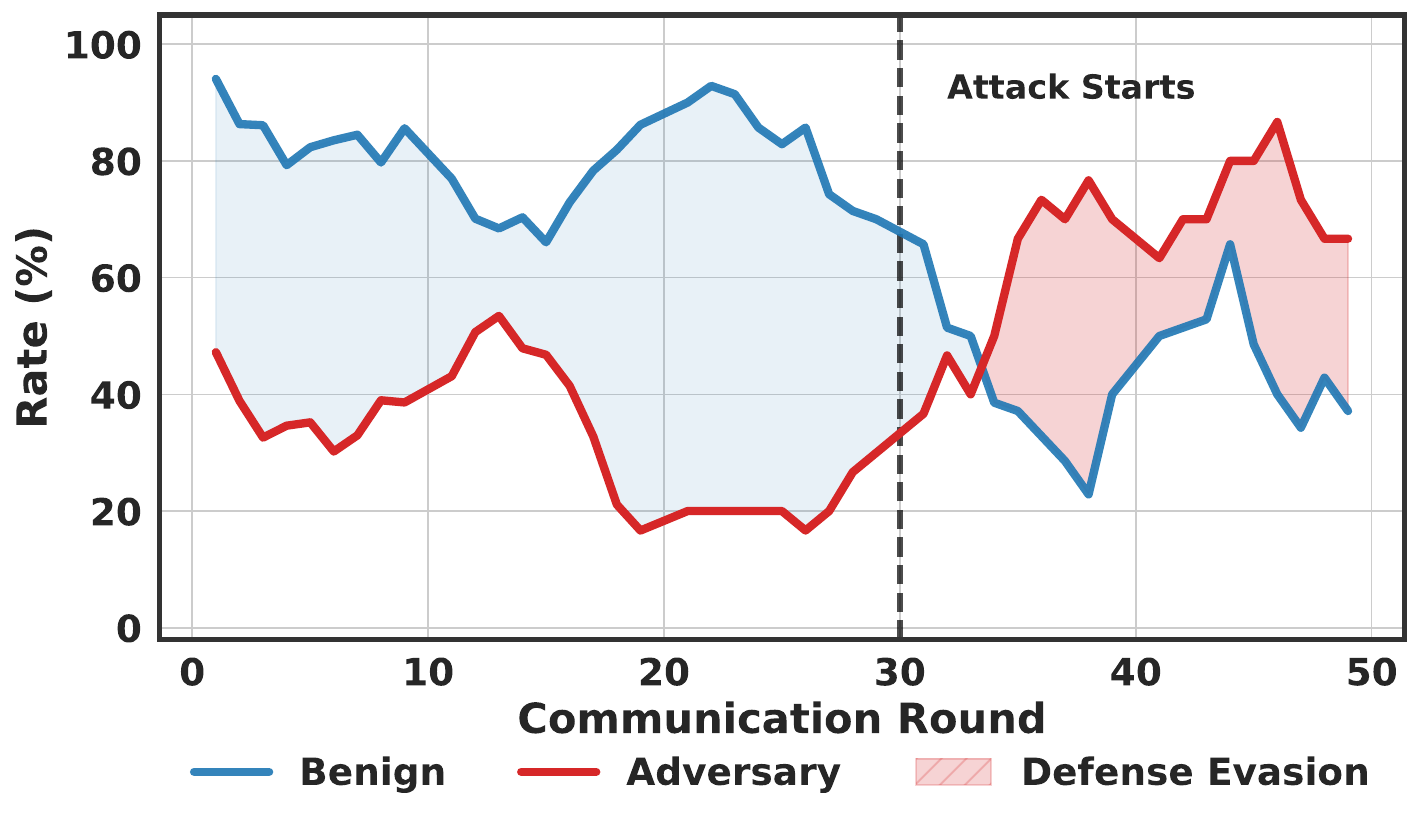}
        \caption{\textbf{Individual Infiltration.} A single bandit learns to capture a high selection rate.}
        \label{fig:individual_infiltration}
    \end{subfigure}
    \vspace{0.2cm} 
    \caption{\textbf{Marketplace Manipulation.} \texttt{MartFL} fails to distinguish ``Quality'' from ``Conformity,'' allowing attackers to capture revenue via artificial density (a) or statistical hygiene (b).}
    \label{fig:attacks_combined}
    \vspace{-0.3cm}
\end{figure}

\noindent\textbf{1. Sybil Drowning.}
We investigate whether a coalition can exploit clustering logic via 
\textbf{Sybil Drowning}, instantiated against \texttt{MartFL} as a 
representative clustering-based filter. We evaluate two coordination 
strategies: \textbf{Mimicry}, where Sybils estimate the benign centroid 
via gray-box probing, and \textbf{Oracle}, an upper-bound strategy with 
full access to the true centroid; the gap between them quantifies the 
cost of imperfect probing. As visualized in 
Figure~\ref{fig:attacks_combined}(a), Sybils form a hyper-dense cluster 
proximal to the estimated ground truth, displacing honest contributions 
from the consensus region. 
This creates a \textbf{Geometric Leverage Effect}: the dense Sybil cluster shifts the global centroid toward itself, causing high-variance honest contributions---which contain novel information---to be reclassified as outliers relative to this shifted mean. The 18-point Mimicry-to-Oracle gap (MSR $80\%$ vs.\ $98\%$) quantifies the market-entry cost of imperfect probing: a coalition with even partial knowledge of the broker's consensus region captures the majority of available revenue, and full knowledge yields near-monopoly---so the attack's economic viability does not require sophisticated coordination.

\noindent\textbf{2. Individual Infiltration.}
Contrary to the assumption that manipulation requires collusion, a single ``Black-Box'' adversary can game the system using a bandit strategy. Learning that the broker penalizes the natural noise of honest sellers, the adversary optimizes for ``Data Hygiene'' by generating low-variance contributions that appear more consistent than honest data. Figure~\ref{fig:attacks_combined}(b) reveals this learning curve:
after a multi-round \textit{Exploration Phase} to probe statistical profiles, the bandit enters an \textit{Exploitation Phase}, achieving a sustained MSR of $\approx60\%$. This proves \texttt{MartFL} rewards \textbf{Statistical Regularity} rather than \textbf{Information Gain}. Crucially, this highlights the economic value of collusion: while the independent bandit pays an exploration cost of tens of rounds, coordinated Sybils capture maximum revenue immediately.

\noindent\textbf{Break-Even Analysis.}
We quantify the attacker's break-even point directly from the bandit's per-round acceptance trace. On \texttt{FLTrust} the adaptive adversary is accepted immediately ($100\%$ selection from the first round), so it incurs no entry cost at all. On \texttt{MartFL}---the only operator imposing meaningful resistance---the bandit must first learn the acceptance region: 
its selection rate climbs over an entry phase of $\approx$$11$--$20$ rounds before stabilizing at $\sim$$60\%$. Treating the entry phase as forgone revenue, the attacker earns $(T-c)\,s\,p$ over $T$ rounds against an opportunity cost of $c\,s\,p$ it would have earned had it been accepted from the outset, so net revenue is $R(T) = (T-2c)\,s\,p$ for entry cost $c$, sustained selection share $s$, and per-round payment $p$; the payment level cancels and break-even occurs at $T^{*} = 2c \approx 31$ rounds, independent of how the marketplace prices contributions.\footnote{Entry cost is definition-sensitive: with three adversaries the per-round selection rate moves in thirds, so we report the entry phase as a range rather than a point estimate. Measured over $n{=}2$ seeds, which agree closely (entry at $11$ and $20$ rounds; sustained rate $\sim$$58\%$ in both), giving $c \approx 15.5$ and $T^{*} \approx 31$. If sellers also bear a per-round participation cost, the horizon lengthens at low payment levels and approaches $2c$ as payment grows.} Against an adaptive gray-box adversary, then, \texttt{MartFL}'s resistance amounts to a bounded entry tax rather than a durable defense---and \texttt{FLTrust} imposes no tax at all.

\begin{insightbox}
Selection rewards statistical conformity, not information gain---coordinated Sybils capture the majority of revenue by mimicking consensus, and even a single adaptive bandit recoups its entry cost within $\approx$31 rounds.
\end{insightbox}

\subsection{Scenario 3: Demand-Side Monopoly Enforcement}
\label{sec:demand_side_bias}
We analyze ``Dual-Sided'' threats where a malicious buyer weaponizes the validation signal ($g_{ref}$) to rig the economy. Both attacks exhibit the same \textbf{architectural asymmetry}: they succeed against \texttt{FLTrust}'s root-of-trust structure but are neutralized by \texttt{MartFL}'s peer-clustering.

\begin{figure}[h]
    \centering
    \begin{subfigure}[b]{0.49\columnwidth}
        \centering
        \includegraphics[width=\linewidth, trim=0.1cm 0.3cm 0.1cm 0.1cm, clip]{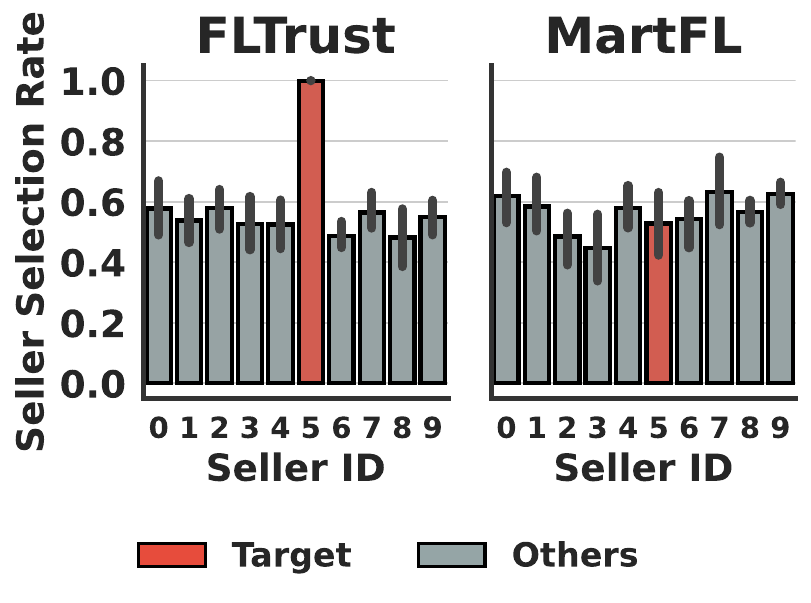}
        \vspace{-0.2cm}
        \caption{\textbf{Monopoly}}
    \end{subfigure}
    \hfill
    \begin{subfigure}[b]{0.49\columnwidth}
        \centering
        \includegraphics[width=\linewidth, trim=0.1cm 0.3cm 0.1cm 0.1cm, clip]{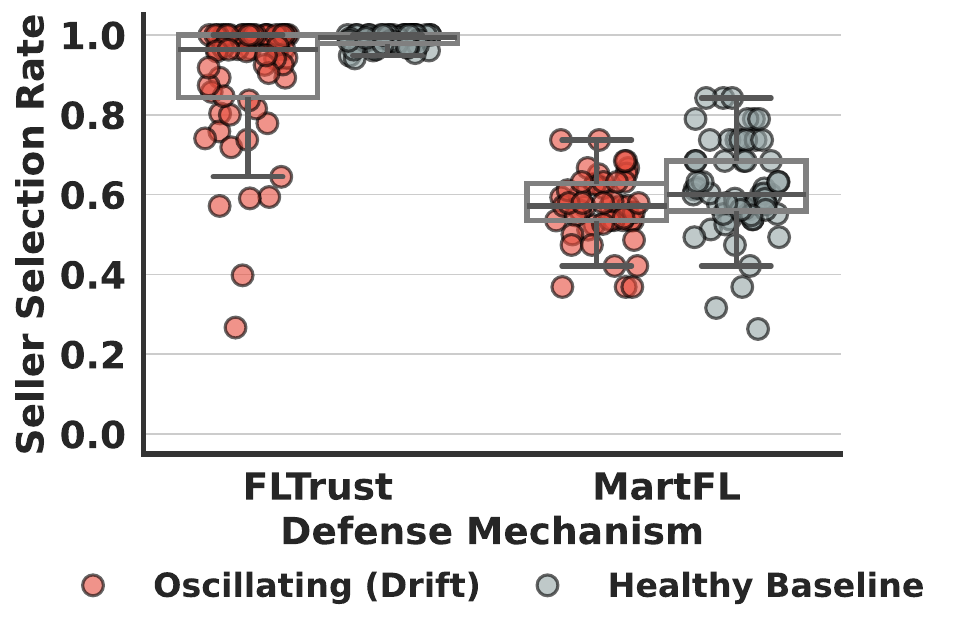}
        \vspace{-0.2cm}
        \caption{\textbf{Market Volatility}}
    \end{subfigure}
    \vspace{-0.2cm}
    \caption{\textbf{Demand-Side Bias.} (a) A buyer forces \texttt{FLTrust} to grant a monopoly. (b) Oscillating attacks cause high variance in stateless filters, preventing market stabilization.}
    \label{fig:buyer_attacks}
    \vspace{-0.5cm} 
\end{figure}

\noindent\textbf{1. Monopoly Enforcement (Geometric Pivoting).} 
The buyer injects noise orthogonal to a favored seller's gradient (Section~\ref{sec:attack_formulation}), making that seller appear optimal relative to the poisoned reference. \texttt{FLTrust} elevates the target to a $100\%$ selection rate while depressing honest sellers into a broad $40$--$70\%$ band (Figure~\ref{fig:buyer_attacks}a), funneling the incentive budget to a colluding seller. \texttt{MartFL} is unaffected---the target is indistinguishable from honest peers---because the orthogonal noise perturbs $g'_{ref}$ without altering the peer-peer distances that drive its decision boundary. Single-reference filters thus create an exploitable single point of failure in a trustless, dual-sided GDM.

\noindent\textbf{2. Inducing Volatility (Dynamic Instability).}
The buyer periodically shifts the validation objective (Section~\ref{sec:attack_formulation}). \texttt{FLTrust}'s per-seller selection rate, tightly concentrated near $1.0$ at baseline, collapses into a wide distribution spanning $0.27$--$1.00$ (Figure~\ref{fig:buyer_attacks}b): unable to maintain stable trust estimates under a shifting reference, the filter delivers inconsistent rewards and burns budget on selection churn rather than new data utility. \texttt{MartFL} again matches its healthy baseline. The root-dependent filter is therefore vulnerable to demand-side manipulation while the peer-comparison defense retains integrity---albeit at the cost of baseline data yield (Section~\ref{sec:baseline_security}). Whether this generalizes across the broader root-based and peer-clustering families is left to future cross-filter evaluation.

\begin{insightbox}
A buyer who controls the query signal can bankrupt competitors without touching a single record: root-of-trust filters inherit the buyer's bias, while peer-clustering filters resist it---at the cost of baseline data yield.
\end{insightbox}

\subsection{Scenario 4: System Scalability \& Provenance Cost (The Broker Evaluation)}
\label{sec:system_scalability}

Beyond security, practical brokers must optimize computational latency and memory overhead. Because GDMs represent saturated ecosystems of persistent, professional sellers, we evaluate absolute system scalability by scaling participation to $N=100$.

\noindent\textbf{Cost, Scale, and Structural Failure.}
Empirical latency contradicts theoretical complexity: \texttt{Multi-Krum}'s $\mathcal{O}(N^2)$ pairwise distances run in near-constant time ($<0.1$s at $N{=}100$) on GPU, while \texttt{SkyMask} ($>180$s) and \texttt{MartFL} ($\approx$$30$s) bottleneck on per-parameter masking and iterative K-means. Memory is the harder constraint---\texttt{FLTrust} and \texttt{MartFL} consume $10$--$13$\,GB VRAM at $N{=}100$, which an SGX-class enclave budget ($128$--$512$\,MB EPC) could hold only through continuous paging (Section~\ref{sec:system_model}). Table~\ref{tab:scalability_selection} shows each filter then failing along its own axis as $N{:}10{\to}100$: \texttt{Multi-Krum}'s fixed rejection fraction produces outright \textbf{Market Paralysis} (BSR $2\%$, near-random accuracy); 
\texttt{MartFL}'s \emph{rates} improve (MSR $43{\to}12$, BSR $64{\to}80$), but the pool grew tenfold, so adversaries drawing payment \emph{rise} from ${\approx}1.3$ to ${\approx}3.6$ per round while ASR holds at $33\%$---the metric improves, the mechanism does not; 
\texttt{FLTrust} exhibits \textbf{Threshold Evasion}, cutting MSR $95{\to}58$ while bypassing adversaries sustain ASR near $70\%$; and \texttt{SkyMask} and \texttt{FedAvg}, lacking per-seller rejection, show no selection response at all. Cost of Convergence confirms this from the budget side: \texttt{SkyMask} spends $271$\,K\,MB of uploads for no security advantage over \texttt{FedAvg} ($279$\,K\,MB), while \texttt{Multi-Krum}'s lower $53$\,K\,MB reflects early divergence rather than efficiency---the run terminates without reaching $\tau_{acc}$.

\begin{table}[h]
\centering
\footnotesize
\setlength{\tabcolsep}{4pt}
\caption{\textbf{Selection Behavior and Cost at Scale (CIFAR-100).} Each cell reports the value at $N{=}10 \rightarrow N{=}100$. CoC: Cost of Convergence (K\,MB uploaded). Full sweep ($N \in \{10,25,50,75,100\}$) in the repository.}
\label{tab:scalability_selection}
\begin{tabular}{l ccccc}
\toprule
\textbf{Filter} & \textbf{BSR} & \textbf{MSR} & \textbf{ASR} & \textbf{Acc} & \textbf{CoC} \\
\midrule
\texttt{FedAvg}     & 100 $\to$ 100 & 100 $\to$ 100 & 80 $\to$ 88 & 47.7 $\to$ 46.2 & 5 $\to$ 279 \\
\texttt{FLTrust}    & 98 $\to$ 98   & 95 $\to$ 58   & 79 $\to$ 68 & 48.9 $\to$ 44.7 & 12 $\to$ 188 \\
\texttt{MartFL}     & 64 $\to$ 80   & 43 $\to$ 12   & 71 $\to$ 33 & 47.4 $\to$ 44.5 & 8 $\to$ 222 \\
\texttt{Multi-Krum} & 20 $\to$ 2    & 54 $\to$ 5    & 88 $\to$ 56 & 18.8 $\to$ 3.2  & 3 $\to$ 53 \\
\texttt{SkyMask}    & --            & --            & 81 $\to$ 88 & 47.5 $\to$ 46.3 & 5 $\to$ 271 \\
\bottomrule
\end{tabular}
\end{table}

\begin{insightbox}
Each filter fails along its own axis at scale: fixed rejection fractions paralyze the market, per-parameter masking exhausts the compute budget, and no filter becomes safer as participation grows once selection is measured in absolute terms rather than as a fraction of a larger pool.
\end{insightbox}

\subsection{Scenario 5: Economic Security \& Incentive Alignment}
\label{sec:valuation}

We audit whether financial incentives align with contribution utility using our \textbf{Shadow Valuation Engine}, comparing realized payment frequency against an estimated utility proxy (KernelSHAP).

\begin{figure}[h]
    \centering
    \includegraphics[width=0.72\columnwidth]{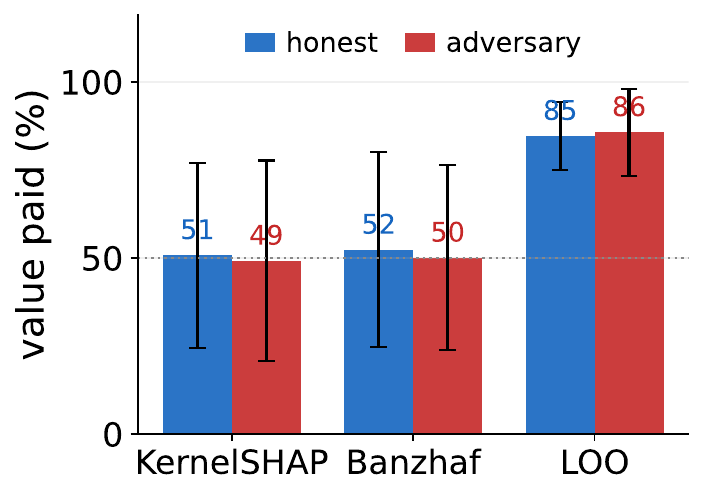}
    \vspace{-0.3cm}
    \caption{\textbf{Valuation--payment misalignment.} Positive value-mass paid to selected sellers on \texttt{MartFL} (CIFAR-100, 5 seeds; error bars = seed std), normalized per method---shares are comparable within a concept, not across. Adversaries capture a share comparable to honest sellers under all three concepts. Influence Functions and the Core do not apply (see text).}
    \label{fig:valuation_gap}
    \vspace{-0.3cm}
\end{figure}

\noindent\textbf{The Valuation Gap.}
Figure~\ref{fig:valuation_gap} exposes a systemic \textbf{Incentive Misalignment}: the
estimated value of a contribution and the payment it receives are decoupled. Under
KernelSHAP, adversarial sellers---despite contributing negatively to the model---capture a
share of total value-mass comparable to honest sellers ($49\%$ vs $51\%$), rather than the
near-zero share their utility warrants. The market thus \textbf{subsidizes malicious
actors} while \textbf{penalizing honest diversity}: high-information-gain sellers, misread
as outliers, are compensated no more than the attackers they are pooled with, and this
adverse selection compounds as underpaid honest sellers exit.
On \texttt{MartFL}---the only filter that selectively rejects, hence the only one where
valuation can disagree with payment---leave-one-out and Banzhaf both assign adversaries a
value share comparable to honest sellers, reproducing KernelSHAP's non-separation. The
magnitude is method-dependent (Banzhaf tracks KernelSHAP near parity; leave-one-out shifts
the level), but the direction is consistent: no estimator we test would, as a payment rule,
correct the adverse selection. Two concepts do not apply. Influence functions assign
near-uniformly negative influence to \emph{all} sellers---defunding the market rather than
misranking it---so they cannot enter a non-negative value-share comparison; and the Core,
evaluated via a Monte-Carlo-sampled least-core LP, is infeasible in every round, being
empty under clustered selection. The mispricing is therefore intrinsic to the valuation
signal, not the payment rule.

\noindent\textbf{Structural Economic Failure Modes.}
The two selective filters fail in opposite ways. \textbf{1. Continuous scoring, binary payment (\texttt{FLTrust}):} the filter suppresses malicious contributions by assigning negligible trust weight, yet retains them in the active pool---so in a selection-as-payment economy the adversary still triggers a payment event, draining the incentive budget while contributing almost nothing to the model. \textbf{2. Market degradation (\texttt{MartFL}):} strict binary exclusion routinely discards high-utility honest contributions, making participation financially unviable; as the multi-round market shows (Section~\ref{sec:results_multiround}), the resulting attrition drives the GDM toward adversarial dominance, with severity governed by how long honest sellers tolerate non-payment.

\begin{figure}[t]
    \centering
    \begin{subfigure}[b]{0.4\columnwidth}
        \centering
        \includegraphics[width=\linewidth]{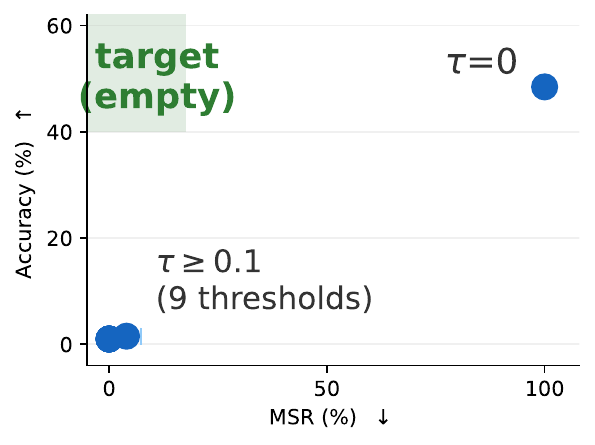}
        \caption{Frontier unreachable.}
        \label{fig:e1_frontier}
    \end{subfigure}
    \hspace{0.05\columnwidth} 
    \begin{subfigure}[b]{0.4\columnwidth}
        \centering
        \includegraphics[width=\linewidth]{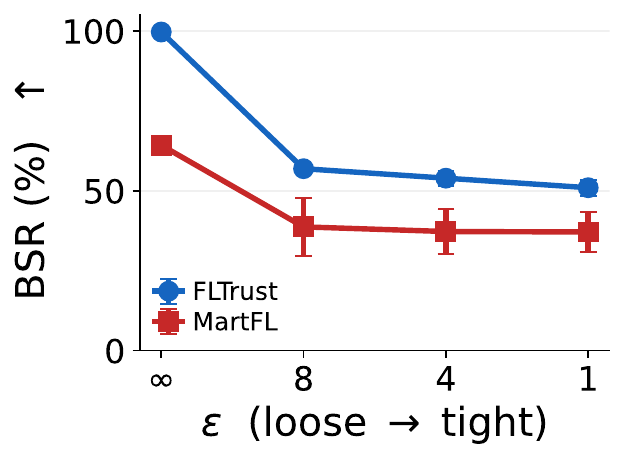}
        \caption{The Privacy Tax.}
        \label{fig:e3_privacy_tax}
    \end{subfigure}
    \caption{\textbf{Two failures that survive the obvious fixes.} (a) Thresholded \texttt{FLTrust} under weight-proportional payment: the target region ($\text{MSR}\leq20\%$, $\text{Acc}\geq40\%$) is empty. (b) As the privacy budget tightens, BSR falls monotonically on both reference-based filters---DP noise is misread as contamination. CIFAR-100, 5 seeds; error bars are seed std.}
    \label{fig:e1_e3_combined}
\end{figure}

\noindent\textbf{Robustness to the Payment Model.}
The failures above are not artifacts of binary payment. Re-running the thresholded-\texttt{FLTrust} sweep under a continuous, weight-proportional payment model (Section~\ref{sec:exp_scenarios})---where each selected seller is paid in proportion to the trust weight the filter already assigns---\emph{no} threshold $\tau \in \{0, 0.1, \dots, 0.9\}$ attains $\text{MSR} \leq 20\%$ with $\text{Acc} \geq 40\%$ (Figure~\ref{fig:e1_frontier}).\footnote{At $\tau=0$ the run completes 78 rounds; for $\tau \geq 0.1$ early stopping halts training at 10 rounds. This is a consequence of the collapse, not a confound: at $\tau \geq 0.1$ the filter selects no sellers ($\text{BSR}=0$), so no aggregation occurs and additional rounds cannot recover accuracy---the model is starved from the first rounds regardless of budget.} The only configuration retaining usable accuracy ($\tau=0$, $48.5\%$) admits every adversary ($\text{MSR}=100\%$), while any threshold that suppresses adversaries ($\text{MSR} \leq 4\%$ at $\tau \geq 0.1$) collapses the model to chance ($\leq 1.5\%$). The frontier is thus a \emph{cliff}, not a tunable trade-off---the Untenable Dilemma persists. Moreover, continuous payment is self-limiting: adversaries capture only $24.6\% \pm 0.6\%$ of total revenue against a $30\%$ population share, a sub-proportional floor; the binary model, if anything, overstates their economic capture. The weighted scheme nonetheless still misprices contribution---by paying proportionally to reference similarity, it rewards camouflaged poisoners aligned with $g_{ref}$ while underpaying legitimate but heterogeneous sellers---the same adverse-selection pattern the Privacy Tax (Section~\ref{sec:results_privacy_tax}) and the valuation gap (Figure~\ref{fig:valuation_gap}) expose.

\begin{insightbox}
The misalignment lies in the valuation signal, not the payment rule---no estimator we test would correct it, and weight-proportional payment leaves the frontier unreachable.
\end{insightbox}

\subsection{The Privacy Tax}
\label{sec:results_privacy_tax}
The submission conjectured that privacy-compliant sellers pay a hidden tax (Section~\ref{sec:validation_paradox}). We test this directly: honest sellers apply a local Gaussian DP mechanism at privacy budget $\epsilon \in \{\infty, 8, 4, 1\}$ ($\epsilon=\infty$: no noise; smaller $\epsilon$: more noise; $\epsilon \in \{1,4,8\}$ calibrated via the Analytic Gaussian Mechanism), while adversaries are never noised. On \emph{both} reference-based filters, honest-seller selection falls monotonically as $\epsilon$ tightens (Figure~\ref{fig:e3_privacy_tax}): \texttt{FLTrust}'s BSR drops from $99.7\%$ to $51.0\%$ ($-48.7$~pp) and \texttt{MartFL}'s from $64.2\%$ to $37.1\%$ ($-27.1$~pp), with no reversal at any step. The filters misread privacy-induced variance as contamination. 
The tax compounds: as honest gradients grow noisier, adversarial selection simultaneously \emph{rises}---\texttt{FLTrust} saturating near $100\%$ and \texttt{MartFL} climbing from $\sim$$42\%$ to $\sim$$67\%$---because the widened honest cluster is easier to hide inside.\footnote{$\epsilon=\infty$ runs $\sim$78 rounds; noised runs $\sim$27 (early stopping fires once accuracy collapses). BSR is a per-round average, so this does not bias it, though accuracy across $\epsilon$ is not compared at equal round budgets. \texttt{MartFL}'s $\epsilon=8$ BSR carries a wider seed spread ($\pm$9.1~pp); its mean remains $>2\sigma$ below the no-noise value, so monotonicity holds.}

\begin{insightbox}
A robust GDM under-selects exactly the sellers a privacy-priced market is designed to pay.
\end{insightbox}

\subsection{Scenario 6: Multi-Round Market Persistence}
\label{sec:results_multiround}
Running the same seller population across sequential tasks (Section~\ref{sec:exp_scenarios}) turns the single-round adverse selection above into a market-composition dynamic, and separates the two selective filters at every setting we test. We vary $k$, the number of consecutive unpaid rounds an honest seller tolerates before exiting, over $\{2, 3, 5\}$. \texttt{FLTrust} loses no honest sellers: its unpaid-streak gap between honest and adversarial sellers is $\approx$$0$ at every $k$, the active market stays at its $\sim$$30\%$ input mix, and nearly all honest sellers survive to the fifth task---because \texttt{FLTrust} admits nearly all contributions (Table~\ref{tab:main_benchmark}), no honest seller is ever starved into exit. \texttt{MartFL} retains fewer honest sellers than \texttt{FLTrust} at every $k$, but the severity is governed by seller patience: when honest sellers are impatient the market collapses outright---$100\%$ adversarial at $k{=}2$ and $95\%$ at $k{=}3$, with essentially none surviving---while at $k{=}5$ it persists in degraded form ($39\%$ adversarial, $5$ of $7$ honest sellers active). The Market Degradation dynamic argued in Section~\ref{sec:valuation} is therefore not an artifact of a single exit threshold, but a monotone function of how long honest sellers tolerate non-payment. Cumulative adversary revenue rises under both filters for opposite reasons: \texttt{MartFL} because the honest market thins, \texttt{FLTrust} because it never excludes adversaries.

\begin{insightbox}
Adverse selection compounds: honest sellers who go unpaid leave, and the market they leave behind is adversarial.
\end{insightbox}

\section{Discussion and Future Directions}
\label{sec:discussion}

Our benchmark reveals a clear reality: state-of-the-art distributed data processing defenses fail to provide economic security in GDMs. Their breakdown exposes severe systemic limitations in software-based data integration, which we distill into critical architectural trade-offs.

\subsection{The Data Validation Paradox \& The Privacy Tax}
\label{sec:validation_paradox}
Validating against a statistically similar reference set enforces homogeneity, yet a GDM exists to acquire heterogeneity: strict validation rejects novel data as outliers, loose validation cannot distinguish it from camouflaged poisoning.

This tension is further exacerbated by the \textbf{Privacy Tax on Honest Sellers}. To satisfy Local Differential Privacy (DP), honest sellers inject statistical noise into their data derivatives. 
Current distance-based integration engines misclassify this DP-induced variance as malicious contamination, actively starving privacy-compliant sellers while easing adversarial camouflage---confirmed empirically in Section~\ref{sec:results_privacy_tax}.

\vspace{1mm}\noindent\textbf{Architectural Requirement: DP-Aware, Utility-Driven Filtering.} 
Future integration mechanisms must move beyond monolithic distance metrics. A viable GDM defense must evaluate the intrinsic structural contribution or information gain per seller, protecting honest diversity while dynamically adjusting outlier thresholds based on a seller's claimed privacy budget ($\epsilon_{DP}$).

\subsection{Provenance Limits \& Demand-Side Vulnerabilities}
\label{sec:beyond_root_data}
Relying on a buyer-provided root dataset creates an exploitable single point of failure: demand-side bias. Manipulating this baseline collapses the entire economic mechanism. Furthermore, our results demonstrate a fundamental limit of software-only filtering: geometric distance metrics cannot reliably distinguish between malicious conformity (Sybils) and benign diversity (heterogeneous data). If a GDM requires absolute, trustless data provenance, architects face a hard engineering bifurcation:
\begin{itemize}[leftmargin=*, nosep]
    \item \textbf{The Cryptographic Overhead Constraint:} Although Zero-Knowledge (ZK) proofs provide computational integrity, they cannot verify source provenance. Securing the actual data source requires Hardware-Rooted Governance (e.g., TEEs). Modern TEE generations~\cite{tee:cheng2024intel} alleviate earlier memory limits, but mandating hardware attestation imposes a compute tax that prices out low-resource sellers.
\item \textbf{The Economic Deterrence Alternative:} If a GDM prioritizes accessibility and throughput over cryptographic hardware, it must abandon real-time filtering in favor of Economic Deterrence, requiring sellers to post financial collateral.
\end{itemize}

\subsection{From Data Filtering to Mechanism Design}
The defense paradigm must therefore shift.
No existing robust-aggregation method treats selection integrity as a first-class objective---they optimize model robustness, not adversary selection rate---so a directly comparable baseline does not yet exist. Constructing one (e.g., the Information-Gain Valuation of Table~\ref{tab:practitioner_guide}, intractable in deep networks today) is the open problem this benchmark motivates.

\vspace{1mm}\noindent\textbf{Architectural Requirement: Trustless Auditing.} 
If a data broker is to operate without vulnerability to demand-side monopolies, it must transition to query-independent valuation (e.g., Decentralized Oracle Networks, or secure multi-party protocols in the vein of secure aggregation~\cite{bonawitz2017practical}) to measure marginal utility improvements without exposing a physical ground-truth dataset.

\vspace{1mm}\noindent\textbf{Architectural Requirement: Collateralized Data Escrow.} 
Concretely, sellers post collateral scaling with their claimed dataset size, slashed when asynchronous audits detect manipulation. Crucially, slashing cannot key on geometric distance, or it re-imposes the Privacy Tax on honest DP noise; audits must instead test whether a contribution violates the variance bound implied by the seller's claimed privacy budget $\epsilon_{DP}$.

\begin{table}[h]
\centering
\vspace{0.2cm}
\caption{\textbf{Proposed Architectural Responses by Threat Landscape.} Based on our findings, we outline candidate directions beyond naive software filtering; each entry identifies the failure mode we empirically demonstrated and the trade-offs of the proposed remedy (which remain future work).}
\label{tab:practitioner_guide}
\resizebox{\columnwidth}{!}{%
\begin{tabular}{llll}
\toprule
\textbf{Market Type} & \textbf{Identified Failure} & \textbf{Recommended Architecture} & \textbf{Trade-offs / Limitations} \\
\midrule
High Diversity & Adverse Selection & Information-Gain Valuation & Computationally intractable \\
(Non-IID Data) & (Rejects honest diversity) & (Measure Information Gain) & (Fragile in deep learning) \\
\midrule
Open Access & Sybil Swarms & Stake-Slashing / Escrow & High financial barrier \\
(Permissionless) & (Crowd-out by Clusters) & (Economic Penalties) & (Risks excluding low-income) \\
\midrule
Competitive & Valuation Rigging & Multi-Party Auditing OR & High protocol latency \\
(Rival Buyers) & (Monopoly Enforcement) & Hardware Attestation (TEEs) & (Severe compute overhead) \\
\bottomrule
\end{tabular}
}
\end{table}

\subsection{Synthesis and Broader Impact}
\label{sec:broader_impact}
Our benchmark indicates that one-size-fits-all filtering is insufficient: security must be tailored to the threat landscape (Table~\ref{tab:practitioner_guide}). Beyond technical security, GDMs introduce societal risks driven by adverse selection and demand-side bias:

\vspace{1mm}\noindent\textbf{1. Economic and Competitive Risks.} 
Distance-based filters demonetize participants whose distributions diverge from the broker's validation baseline; at scale this risks systematically marginalizing exactly the sellers a marketplace most needs, an effect warranting dedicated fairness evaluation. Separately, resourceful buyers can manipulate valuation to bankrupt competitors, which mandated audit trails for valuation algorithms would help deter.

\vspace{1mm}\noindent\textbf{2. Responsible Disclosure.} The attacks introduced here target deployed filtering mechanisms---not software vulnerabilities---and require no privileged access beyond normal GDM participation. We release this benchmark as design guidance for next-generation GDM architectures rather than operational attack recipes.

\section{Conclusion}
\label{sec:conclusion}
This paper introduces the first benchmark dedicated to auditing 
the Mechanism Integrity of Gradient Data Marketplaces (GDMs). 
We empirically demonstrate that state-of-the-art integration 
operators fail to secure GDM viability, suffering from a systemic 
valuation gap where honest data heterogeneity is penalized as 
outlier behavior while malicious conformity is subsidized. 
Furthermore, we identify demand-side vulnerabilities where a 
buyer can weaponize the query signal to enforce data monopolies. 
These failures are structural rather than artifacts of 
our accounting: they persist under continuous, weight-proportional 
payment, compound across repeated market rounds as underpaid honest 
sellers exit, and worsen when honest sellers adopt differential 
privacy---whose noise the filters misread as contamination. As the community transitions toward decentralized data ecosystems, 
our open-source framework provides the critical infrastructure 
necessary to evaluate and design economically secure, 
utility-driven data integration engines.

\begin{acks}
This work was supported in part by NSF grants 2442825 and 2613659. We thank the anonymous reviewers for their feedback and suggestions.
\end{acks}



\end{document}